\documentclass[lettersize,journal]{IEEEtran}
\usepackage{amsmath,amsfonts}
\usepackage[noend]{algorithmic}
\usepackage{algorithm}
\usepackage{array}
\usepackage[caption=false,font=normalsize,labelfont=sf,textfont=sf]{subfig}
\usepackage{textcomp}
\usepackage{stfloats}
\usepackage{url}
\usepackage{verbatim}
\usepackage{graphicx}
\usepackage{cite}
\usepackage{booktabs}
\usepackage{multirow}
\usepackage{makecell}
\usepackage{capt-of}
\usepackage{array}
\usepackage{ragged2e}
\usepackage{booktabs}
\newcolumntype{L}[1]{>{\RaggedRight\arraybackslash}p{#1}}
\usepackage[table]{xcolor}
\usepackage{tikz}
\usepackage{pifont}
\usepackage[colorlinks=true,linkcolor=black,citecolor=black,urlcolor=magenta]{hyperref}
\definecolor{projectgreen}{RGB}{27,122,28}
\newcommand{\cmark}{\textcolor{green!60!black}{\ding{51}}}
\newcommand{\xmark}{\textcolor{red}{\ding{55}}}
\newcommand{\algcomment}[1]{\hfill {\scriptsize\textcolor{gray}{// #1}}}

\begin{document}

\title{\vspace{-1.5mm}RL$^2$\text{-}VLA: Adaptive RL Latent Compositional Steering with\\\vspace{-0.5mm}
Test-Time Scaling for Vision-Language-Action Models}

\author{
    \vspace{-0.5mm}
    Derek Ming Siang Tan$^{1,3\dagger}$\quad
    Shailesh Shailesh$^{1\dagger}$\quad
    Srikrishna Iyer$^{3}$\quad
    William Wei Jie Teo$^{1,3}$\\
    Yuanliang Ju$^{2}$\quad
    Qiao Gu$^{2}$\quad
    Guillaume Sartoretti$^{1}$\\
    {\vspace{+1mm}\normalsize
        $^{1}$National University of Singapore\quad
        $^{2}$University of Toronto\quad
        $^{3}$Singapore Technologies Engineering\quad
    }\\
    {\vspace{+1mm}
        \href{https://rl2-vla.github.io}{\textcolor{projectgreen}{\rmfamily\mdseries https://rl2-vla.github.io}}
    }

}

\IEEEaftertitletext{
\vspace{-1.47cm}
\begin{center}
\refstepcounter{figure}
\includegraphics[width=\textwidth]{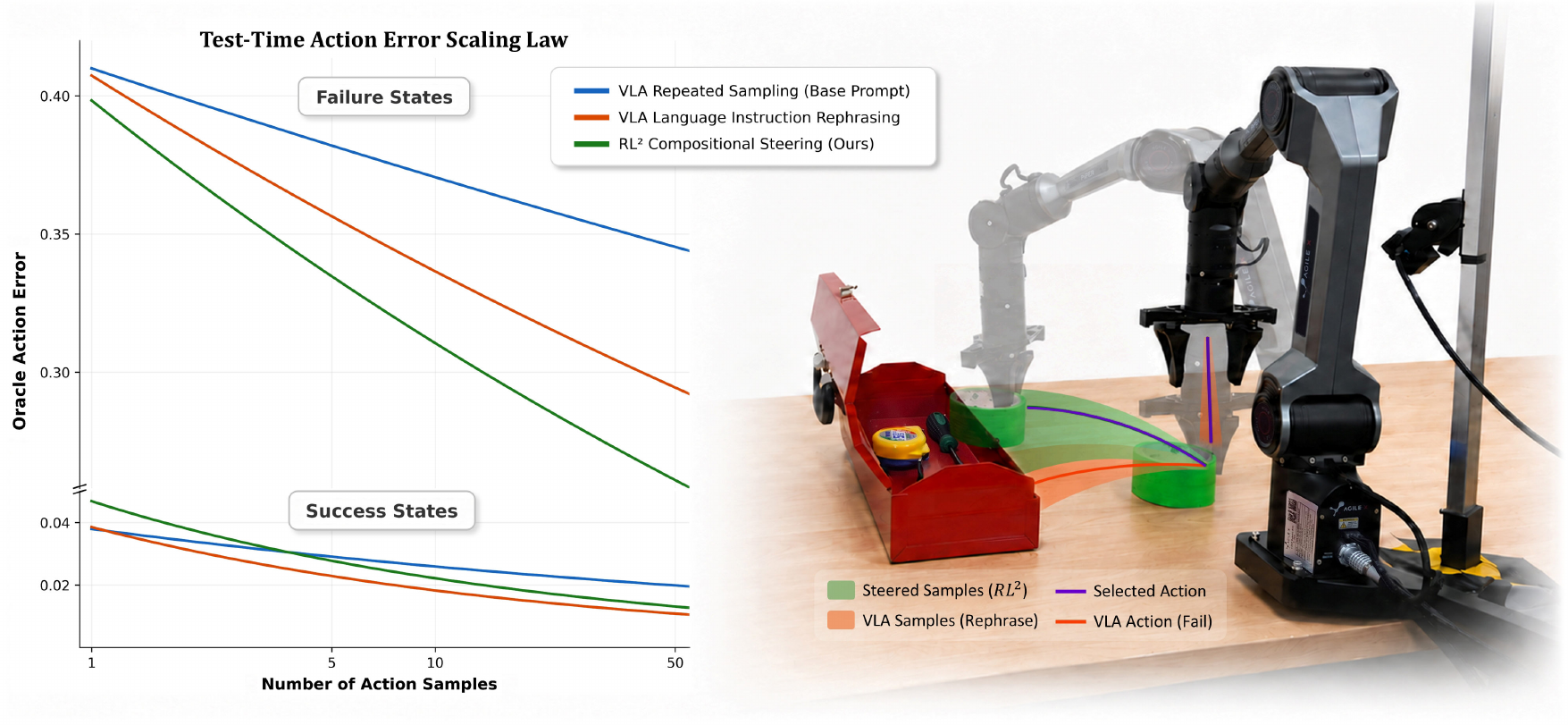}
\label{fig:poster_fig}
\par\vspace{-1.0em}
\begin{minipage}{\textwidth}
\vspace{-0.3cm}
\normalfont\footnotesize\noindent Fig.~\thefigure.\nobreakspace\nobreakspace 
\textbf{Overview:} $RL^2$ improves VLA test-time scaling by adaptively applying RL compositional steering when the base VLA is likely to fail, particularly in OOD settings, without modifying the pretrained VLA.
\textbf{(Left)} Our scaling laws indicate that RL compositional steering (green) reduces action error most effectively under failure states, but can unnecessarily perturb already-accurate success states, motivating adaptive steering. 
\textbf{(Right)} During deployment, $RL^2$ initially follows the base VLA (orange) to accurately grasp the tape, then performs compositional steering (green) after impending failure is detected, generating diverse action candidates that lift the tape sufficiently high for successful placement into the toolbox (lighting and viewpoint slightly enhanced via GenAI).
\end{minipage}
\end{center}
\vspace{1mm}
}

\markboth{}{}

\maketitle

\renewcommand{\footnoterule}{
  \kern -3pt
  \hrule width 0.4\columnwidth height 0.4pt
  \kern 2.6pt
}

\begingroup
\renewcommand{\thefootnote}{}
\footnotetext{\textsuperscript{\(\dagger\)} denotes equal contribution in no particular order}
\addtocounter{footnote}{-1}
\endgroup

\begin{abstract}
Despite the impressive visuomotor capabilities enabled by Vision-Language-Action (VLA) models, their performance often degrades on challenging and out-of-domain tasks. 
Recent test-time steering and scaling methods improve performance without extensive data collection and retraining, but action samples often remain concentrated around similar behaviors and therefore inherit correlated failure modes. 
Moreover, existing methods apply the same intervention strategy at every timestep, regardless of whether the base policy is already likely to succeed. 
To address these limitations, we introduce $RL^2$, an adaptive inference-time steering framework that leverages \underline{R}einforcement \underline{L}earning on VLA \underline{L}atents. 
First, we train a lightweight offline RL policy conditioned on expressive latents extracted from the VLA action expert and compose its flow velocity with that of the frozen VLA during inference.
This compositional steering strategy combines the behavioral priors of large-scale imitation learning with the action diversity induced by offline RL beyond dominant demonstration modes.
We further discover that inference-time steering follows fundamentally different scaling laws under success and failure states, revealing that action diversity is most beneficial when the base VLA is likely to fail, but can unnecessarily perturb already-accurate actions when success is likely.
Building on this insight, $RL^2$ activates compositional steering only when failure is predicted.
Across the SIMPLER and PolaRiS benchmarks, $RL^2$ improves success rates by up to +17.3\% in out-of-domain settings,
while ablations and scaling studies demonstrate the importance of latent representations and RL training. 
Finally, real-world experiments demonstrate that these gains transfer beyond simulation, establishing $RL^2$ as a practical and modular steering framework for VLA deployment.
\end{abstract}

\begin{figure*}[!t]
\vspace{-3mm}
\centering
\includegraphics[width=\textwidth]{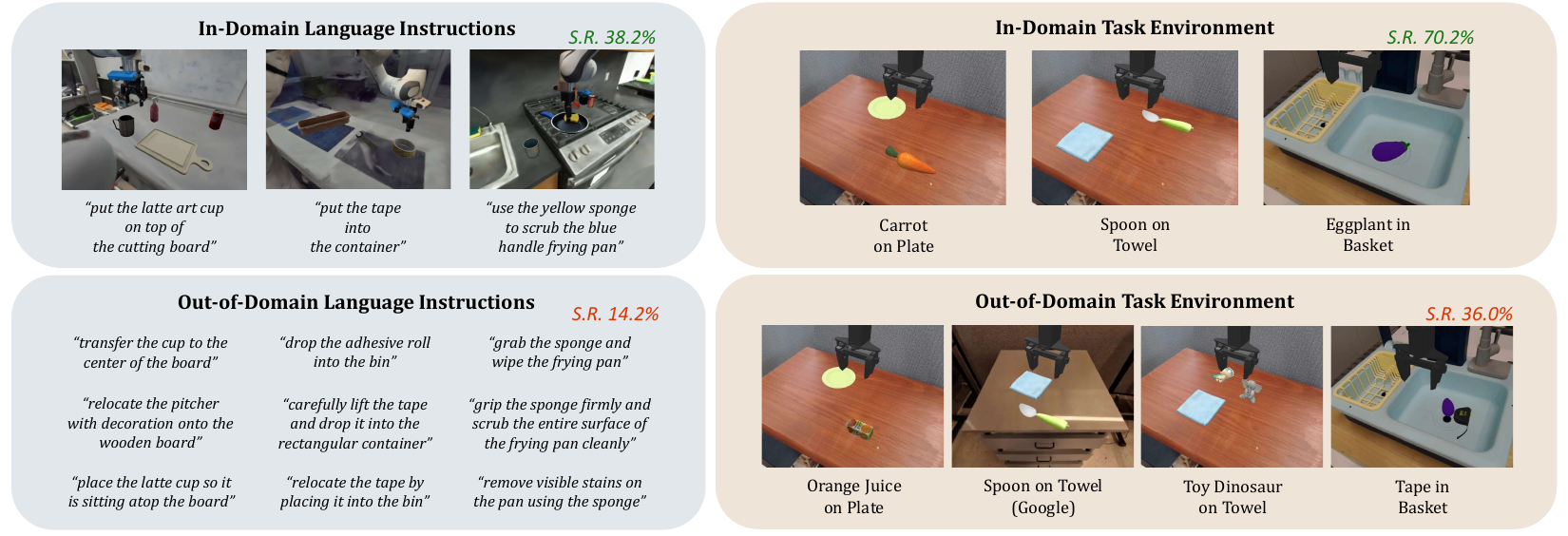}   
\vspace{-7mm}
\caption{\textbf{Types of In-domain and OOD Tasks:} While VLAs often perform well for in-domain tasks seen during training, their success rates tend to deteriorate when encountering unseen language instructions (left, $\text{38.2\%}\rightarrow\text{14.2\%}$)~\cite{fang2025intact} or unseen task environments (right, $\text{70.2\%}\rightarrow\text{36.0\%}$)~\cite{fan2026interleavevla}. Recent methods focus on inference-time steering for VLAs to achieve better generalization to these OOD tasks without extensive data collection and expensive retraining.}
\label{fig:iid_vs_ood_expt_examples}
\vspace{-0.4cm}
\end{figure*}

\vspace{-5mm}
\section{Introduction}
\IEEEPARstart{G}{eneralist} policies have emerged as a promising paradigm for general-purpose robotics, with Vision-Language-Action models (VLAs) among the most prominent examples.
By leveraging large-scale robot demonstrations and strong priors from Vision-Language Models (VLMs), VLAs have demonstrated increasingly robust and generalizable performance across a diverse range of manipulation tasks~\cite{kim2024openvla,black2026pi0,intelligence2025pi05,lee2025molmoact}.
Significant effort has been devoted to improving these models during pretraining through data scaling~\cite{oxe2023oxe} and architectural design~\cite{black2026pi0}, and during post-training through preference alignment~\cite{zhang2025grape} and chain-of-thought reasoning~\cite{zawalski2024ecot,duan2025fastecot}.
More recently, attention has shifted toward improving VLAs during deployment.
These approaches leverage additional computation at test time without requiring extensive data collection and retraining.
They aim to improve robustness especially in out-of-domain (OOD) settings where VLA success rates often deteriorate~\cite{fang2025intact,fan2026interleavevla}, as seen in Fig.~\ref{fig:iid_vs_ood_expt_examples} (e.g. $\text{70.2\%}\rightarrow\text{36.0\%}$).
Existing methods broadly follow two parallel research directions: inference-time steering and test-time scaling.

\textbf{Inference-time steering} improves pretrained VLAs at deployment and can be broadly categorized into two directions. 
The first performs \emph{discrete action selection}~\cite{nakamoto2024steering,kwok2025robomonkey,kwok2026scaling,wu2025foresight,wu2025say}, sampling multiple candidate actions and selecting among them using external verifiers. 
While preserving the policy's learned action distribution, these methods are limited by sample diversity, restricting adaptation to OOD scenarios. 
The second employs \emph{differentiable steering}~\cite{du2025dynaguide,liu2026vls,li2026pilot} to guide action generation through VLM-derived scoring functions or differentiable metrics computed from predicted future outcomes via world models. 
Although capable of steering policies beyond their original action distribution, these approaches are limited by imperfect physical grounding in pretrained VLMs and the difficulty of learning accurate world models~\cite{liu2026vls}.
In parallel, recent works have explored \textbf{test-time scaling} for VLAs to enhance inference-time steering.
Early approaches repeatedly sample actions from the same VLA policy and rely on external verifiers to select the best candidate~\cite{nakamoto2024steering,kwok2025robomonkey}, while more recent methods generate diverse action samples from rephrased language instructions~\cite{kwok2026scaling}. 
Despite these advances, all generated samples ultimately originate from the same policy and may therefore inherit correlated biases and failure modes. 
More fundamentally, existing steering and scaling approaches apply a fixed intervention strategy 
and do not distinguish between situations where the base VLA is already likely to succeed and those where intervention may be necessary.

Humans, however, exhibit markedly different behavior:
\textbf{When faced with uncertainty, humans actively broaden the diversity of alternatives they consider.}
For example, when reaching for a mug on an uncluttered table, we typically grasp it directly without deliberating on alternatives.
However, if the mug is partially occluded, we may approach it from another side or first remove the obstacle in the way.
Thus, diverse alternatives are most useful when the default behavior is likely to fail, but may introduce unnecessary perturbations when the base solution is likely to succeed.
This leads to the question:

\vspace{0.5em}
\begin{quote}
\noindent\vrule width 2pt\hspace{0.75em}%
\begin{minipage}{0.93\linewidth}
\emph{\textbf{How} and \textbf{when} should we steer pretrained VLAs toward diverse action candidates to enhance robustness in challenging / OOD scenarios?}
\end{minipage}
\end{quote}
\vspace{0.5em}

We address this with $RL^2$, our proposed test-time steering framework that leverages \underline{R}einforcement \underline{L}earning on VLA \underline{L}atents for adaptive compositional steering.
To answer the question of \textbf{\textit{how to steer}}, we introduce a lightweight RL-based flow-matching policy conditioned on expressive latents extracted from the VLA action expert.
It is trained on the same offline finetuning dataset used by the VLA and introduces minimal computational overhead.
Thereafter, we use the flow velocity of this RL policy to steer the flow velocity of the VLA action expert via composition (i.e., weighted average of velocities)~\cite{cao2026compose}.
We term this process \textbf{\textit{compositional steering}}, which is particularly effective because it combines the benefits of imitation learning (IL) and RL.
Specifically, the VLA contributes strong behavioral priors learned from large-scale but inherently imperfect demonstrations, while offline RL encourages the discovery of alternative high-value behaviors and induces greater action diversity beyond dominant demonstration modes (action sample illustrations in Fig.~\ref{fig:rl2_viz_cover_samples}).

To answer the question of \textbf{\textit{when to steer}}, we extend the test-time scaling laws established by existing works~\cite{kwok2025robomonkey,kwok2026scaling} to steering approaches under success and failure states.
Assuming an oracle verifier, we find that the relationship between normalized action error and the number of generated action samples follows an exponential power law.
Our scaling law analysis reveals a key insight that aligns well with human intuition: \textbf{the diversity induced by compositional steering is most useful when the base VLA is likely to fail}. 
More specifically, compositional steering can avoid dominant failure modes, but may unnecessarily degrade already-accurate actions when success is likely.
Motivated by this observation, we incorporate SAFE~\cite{gu2026safe}, a failure detector for VLAs, into $RL^2$ to adaptively enable compositional steering during failure and to fall back on the base VLA during success (Fig.~\ref{fig:real_tape_in_toolbox}).
To the best of our knowledge, we are the first work to establish distinct test-time scaling laws that compare different steering approaches separately for success and failure states.

We evaluate $RL^2$ on both the SIMPLER~\cite{li24simpler} and PolaRiS~\cite{jain2025polaris} benchmarks, covering diverse manipulation tasks under in-domain and OOD settings using different verifiers. 
Across these benchmarks, $RL^2$ consistently improves success rates by up to +17.3\% over the strongest baselines for OOD tasks. 
We further perform ablation studies to highlight the effectiveness of using VLA latents and RL training, as well as additional studies that demonstrate favorable scaling with both sample quantity and action diversity.
Finally, we validate our approach on-hardware on a PiperX manipulator, where compositional steering improves success rate by an average of +17.5\% over our strongest baseline, demonstrating the effectiveness and modularity of $RL^2$ for real-world deployment.
Our main contributions are summarized as follows:

\begin{figure*}[!t]
\vspace{-3mm}  
\centering
\includegraphics[width=0.99\textwidth]{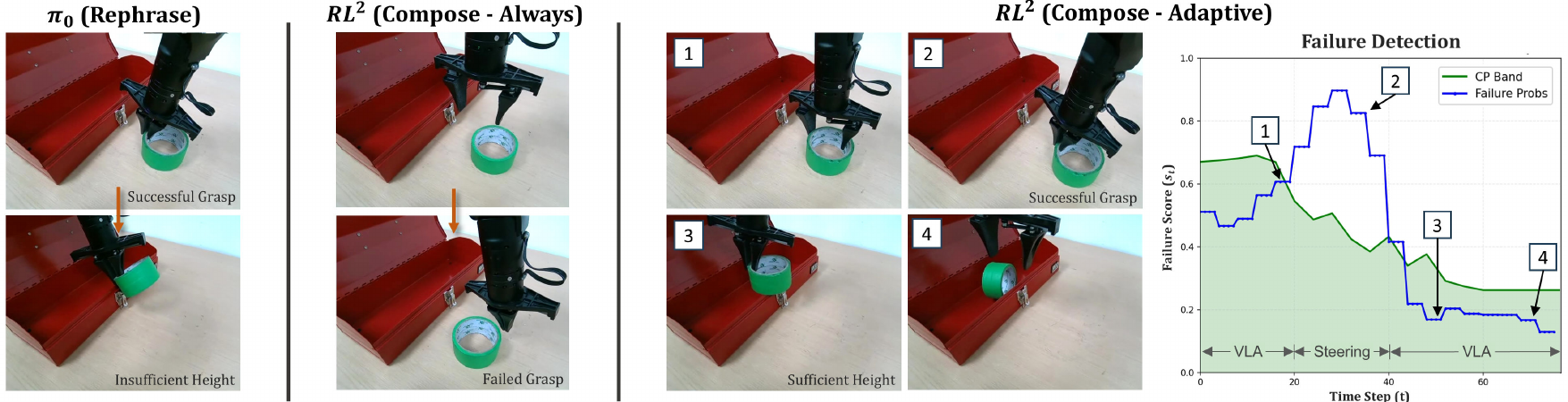}
\vspace{-4mm}
\caption{\textbf{Real-Robot Experiment for \textit{Tape in Toolbox} Task (OOD Environment):} 
This task is challenging for the base VLA ($\pi_0$~\cite{black2026pi0}) because the tape was never seen during training. 
\textbf{(Left)} \textit{Rephrase} correctly grasps the tape but often collides with the side of the toolbox. 
\textbf{(Middle)} Non-adaptive $RL^2$ inaccurately approaches the tape, as it unnecessarily steers already-accurate VLA action samples.
\textbf{(Right)} Adaptive $RL^2$ relies on accurate VLA samples to approach the tape, then steers actions to lift the tape high enough to avoid collision, when failure is preemptively detected via Conformal Prediction (CP)~\cite{gu2026safe}.}
\label{fig:real_tape_in_toolbox}
\vspace{-0.5cm}
\end{figure*}

\begin{itemize}
    \item We demonstrate that a lightweight offline RL policy conditioned on VLA latents and trained on the same VLA finetuning dataset can substantially improve downstream performance through compositional steering.
    \item We establish distinct test-time scaling laws for VLA steering approaches under success and failure states, revealing that RL compositional steering is beneficial during failure but degrades performance during successful states.
    \item We highlight the modularity of our lightweight $RL^2$ framework through extensive validation across diverse VLAs, verifiers, and manipulation benchmarks.
\end{itemize}

\section{Related Works}
\subsection{Imitation vs Reinforcement Learning for VLAs}
Many Vision-Language-Action (VLA) models are trained via behavior cloning on large-scale offline datasets~\cite{kim2024openvla,black2026pi0,intelligence2025pi05,lee2025molmoact}. 
While this paradigm has endowed policies with impressive generalization, they remain constrained by behaviors in the demonstration data. 
RL overcomes this limitation by enabling policies to discover high-value behaviors beyond demonstrations. 
One line of work uses RL to train entire VLA models end-to-end, such as RECAP ($\pi^{*}_{0.6}$)~\cite{intelligence2025pi06} and SimpleVLA-RL~\cite{li2026simplevlarl}.
Although effective, updating large foundation models remains computationally expensive and often requires careful optimization to maintain training stability.

\subsection{Inference-Time Steering}
Inference-time steering methods aim to improve pretrained VLAs at deployment without updating the base model. 

\textbf{Discrete Action Selection:} 
These methods generate multiple candidate actions and rely on external verifiers to select the most promising one. 
V-GPS~\cite{nakamoto2024steering} employs a learned value function to rank policy samples, while RoboMonkey~\cite{kwok2025robomonkey} and CoVer~\cite{kwok2026scaling} use significantly larger VLM-based verifiers trained through preference learning and contrastive learning, respectively. 
FOREWARN~\cite{wu2025foresight} and Do What You Say~\cite{wu2025say} further augment verification with world-model rollouts.
While these approaches preserve the strong behavioral priors learned from large-scale demonstrations, their ability to adapt to desired behaviors outside of this distribution remains limited.

\textbf{Differentiable Steering:}
Rather than selecting among a fixed set of action candidates, differentiable steering methods directly influence the action-generation process using continuous guidance signals. 
DynaGuide~\cite{du2025dynaguide} leverages a learned world model to predict future latent outcomes and steer actions according to guidance objectives, while VLS~\cite{liu2026vls} and VLA-Pilot~\cite{li2026pilot} employ VLM-derived scoring functions to guide the denoising process. 
Although these methods can steer policies beyond their original action distribution, they depend heavily on external guidance models; pretrained VLMs may lack physical grounding, while world models can introduce prediction errors and substantial inference cost~\cite{liu2026vls}.

\textbf{Direct Action Refinement:} 
Another line of work improves VLAs through lightweight auxiliary policies.
ConRFT~\cite{chen2025conrft} freezes the VLA encoder and adapts an action head using consistency-based RL training objectives.
Policy Decorator~\cite{yuan2025decorator} and PLD~\cite{xiao2026pld} each learn a residual policy that outputs corrective offsets, while DSRL~\cite{wagenmaker2025steering} uses RL to steer noise inputs of diffusion policies to enable efficient online adaptation. 
Similarly, RLT~\cite{xu2026rltoken} learns compact tokens used by RL policies for rapid online learning. 
While effective, these methods often require online interaction with real or closely-simulated environments, increasing deployment cost.

Our approach steers the VLA using latent RL composition while retaining verifier-based selection.
This combines the robustness of discrete action selection with the ability of differentiable steering to explore beyond the policy distribution.

\subsection{Test-Time Scaling}
Test-time scaling (TTS) aims to improve performance by allocating additional computation during inference.
The first line of work improves inference-time steering methods by applying TTS to action generation. 
RoboMonkey~\cite{kwok2025robomonkey} characterizes the first inference-time scaling law for VLAs through repeated action sampling and verifier-based selection, while CoVer~\cite{kwok2026scaling} promotes additional diversity by sampling actions conditioned on multiple language instruction rephrases. 
However, additional samples drawn from the same VLA policy may simply generate diverse variants of the same failure mode. 
Another line of work uses embodied chain-of-thought reasoning~\cite{zawalski2024ecot,duan2025fastecot} during inference to decompose long-horizon tasks into intermediate plans before generating robot actions.
Our approach extends TTS for action generation by characterizing distinct scaling behaviors for steering approaches under success and failure states, and introduces latent RL steering to diversify candidate actions only during failure when it helps most.

\subsection{Adaptive Inference for VLAs}
Recent works on adaptive inference for VLAs aim to dynamically adjust the inference process according to the difficulty of the current state.
Existing approaches primarily differ in the signals used to trigger adaptation. 
OneTwoVLA~\cite{lin2026onetwovla} learns routing tokens that determine whether the model should reason or act directly, while Recurrent-Depth VLA~\cite{tur2026rdvla} dynamically adjusts latent refinement depth according to action consistency. 
Under high uncertainty, VLA-ATTC~\cite{li2026vlaattc} invokes a deliberative action-selection process, whereas SCALE~\cite{choi2026scale} broadens visual attention and action diversity.
In contrast, we adopt a failure-centric perspective using SAFE~\cite{gu2026safe}, a state-of-the-art failure detector which outperforms uncertainty-based methods.
Similar to SCALE, our goal is to diversify behaviors when failures are anticipated; however, rather than broadening exploration by adjusting the temperature hyperparameter of the same base policy, we employ a separately-trained lightweight RL policy to compositionally steer VLA actions.

\section{Preliminaries}

\subsection{Imitation and Reinforcement Learning}
VLA models are typically trained via imitation learning on large-scale robot demonstration datasets $\mathcal{D}=\{o_i, a_i, \ell_i\}_{i=1}^{N}$, where $o_i$ denotes the robot observation, $a_i$ the corresponding action, and $\ell_i$ the task language instruction. 
Depending on the VLA architecture, observations may consist of one or more images together with robot proprioception, while the policy typically predicts an action chunk $a_{t:t+H}$ over a horizon $H$. 
The policy $\pi_\theta$ is trained by maximizing the log-likelihood of action chunks, $\max_{\theta}\mathbb{E}_{(a_{t:t+H}, o_t, \ell_t)\sim\mathcal{D}}\left[\log\left(\pi_\theta(a_{t:t+H}\mid o_t, \ell_t)\right)\right]$.
This objective encourages the policy to imitate the action distribution present in large-scale offline datasets.

In addition, we consider the standard RL formulation, modeled as a partially observable Markov Decision Process (POMDP). 
At each decision step, the agent selects and takes an action based on its policy $\pi_\theta(a_{t:t+H}|o_t,\ell_t)$. 
The overall goal of RL is to maximize the expected cumulative return $ R = \mathbb{E}_{a_{t:t+H} \sim \pi(\cdot \mid o_t,\ell_t)}\left[\sum_{t=1}^{T} \gamma^{t-1} r^t \right].$

\subsection{Classifier Guidance for Flow-Matching Policies}
Classifier guidance~\cite{dhariwal2021diffusion} provides a mechanism for steering pretrained generalist policies during inference. 
For a VLA with a flow matching head, action generation is governed by a velocity field $v(o,a,\ell,k)$ that progressively transports an initial noise sample toward a valid action trajectory.
The flow-matching process models an Ordinary Differential Equation (ODE), where $k$ denotes the flow-matching progress.
Given a differentiable guidance function $g(o,a,\ell)$ encoding a desired steering objective, the velocity field can be modified as:
\begin{equation}
\hat{v} = v\left(o_t, a_{t:t+T}, \ell_t, k\right) + \lambda \cdot g\left(o_t, a_{t:t+T}, \ell_t\right).
\end{equation}
where $\lambda$ controls the strength of guidance. 
Intuitively, the guidance function biases the flow matching process toward action trajectories that better satisfy a desired objective while preserving the underlying policy prior.

\begin{figure*}[!t]
\centering
\includegraphics[width=\textwidth]{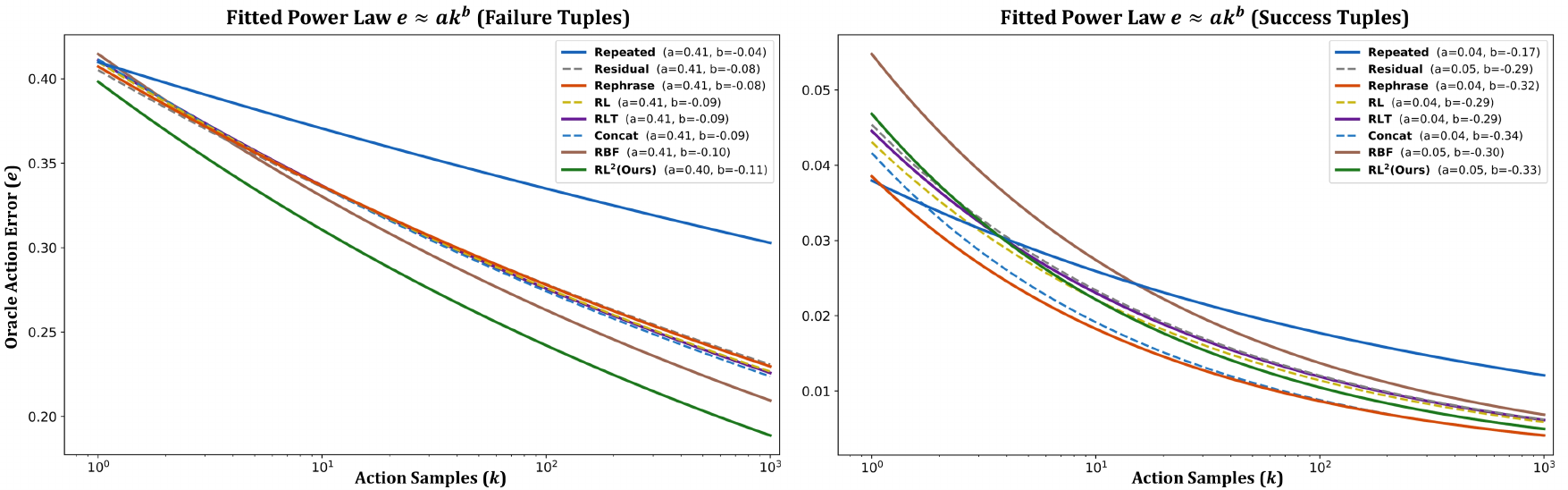}
\vspace{-6mm}
\caption{\textbf{Test-time Scaling Laws (Steering Improves Failure States but Degrades Success States):} 
We observe that the action error between ground truth and predicted VLA actions ($\pi_0$~\cite{black2026pi0}) consistently decreases as we scale the number of action samples across different offline RL and baseline steering approaches. 
Most notably, compositional steering ($RL^2$) and \textit{RBF}~\cite{jeon2026tree} exhibit the best scaling property due to their ability to generate diverse yet high-quality action samples during failure. However, they perform poorly compared to baseline approaches during success states, prompting the need to apply steering adaptively.}
\label{fig:inference_time_scaling_plots}
\vspace{-0.4cm}
\end{figure*}

\section{Inference-Time Scaling Law}
\label{sec:nrmse_inference_scaling_law}
We extend the test-time scaling laws established by existing works~\cite{kwok2025robomonkey,kwok2026scaling} to various steering approaches under success and failure states, with the goal of characterizing how generated sample count affects action error across methods and states.
Using $\pi_0$~\cite{black2026pi0} as our base VLA, we first generate the policy actions for each tuple $(s, a^*, I)$ within the validation set of the large-scale BridgeV2 dataset~\cite{walke2023bridgedata}.
We then compute the normalized root-mean-squared error (NRMSE) between policy and ground truth actions, and extract the top and bottom 1,024 tuples based on NRMSE to define the failure and success tuple sets respectively.
The NRMSE formula per tuple is as follows:
\begin{equation}
\mathrm{NRMSE}
=\sqrt{\frac{1}{N}\sum_{i=1}^{N}
\left(
\frac{a_i-a^{*}_i}{\max_T a_i - \min_T a^{*}_i}
\right)^2}.
\label{eq:nrmse}
\end{equation}
where $a$ denotes the policy action, $a^*$ the ground truth actions, $N$ the number of flattened action dimensions, and $T$ tuples used from BridgeV2 dataset.
Note that while there exist failure detection methods that can be trained using only successful trajectories from offline datasets, our approach offers a simpler, non-training-based alternative to determine policy failure since we have access to ground truth actions in BridgeV2.
 
We evaluate the following baseline steering approaches by taking the average NRMSE across each of the tuple sets: 
\begin{enumerate}
    \item \textbf{\textit{Repeated}~\cite{kwok2025robomonkey}}: actions are repeatedly sampled from the VLA $\pi(a \mid o, l)$ using the same language instruction $l$.
    \item \textbf{\textit{Rephrase}~\cite{kwok2026scaling}}: actions are sampled from the VLA $\pi(a \mid o, l)$ where input $l$ are VLM-generated rephrases.
    \item \textbf{\textit{RBF}~\cite{liu2026vls}}: actions are sampled similarly to \textit{Rephrase}, but the first $n$ flow matching steps use Radial Basis Function (RBF) repulsion~\cite{jeon2026tree} to promote diversity. 
\end{enumerate}

We further evaluate other offline RL steering approaches, conditioned on rephrases and latents $e$ from the VLA action expert (more details regarding $e$ in Appendix~\ref{app:vla_feature_extraction}).
\begin{enumerate}
\setcounter{enumi}{3} 
    \item \textbf{\textit{RL}~\cite{li2026qam}}: actions are repeatedly sampled from an RL flow-matching policy $\pi(a \mid e)$ trained using \textit{Q-learning with Adjoint Matching} (QAM).
    \item \textbf{\textit{Concat}}: actions are sampled equally from both the base VLA and the RL flow-matching policy.
    \item \textbf{\textit{Residual}}~\cite{yuan2025decorator}: residual actions are sampled from an offline RL policy $\pi_{res}(a \mid e, a_{VLA})$, and are then added to steer the base VLA actions. We adopt the implementation from~\cite{yuan2025decorator}, but trained using offline RL.
    \item \textbf{\textit{RLT}}~\cite{xu2026rltoken}: actions are sampled from an offline RL policy $\pi(a \mid e, a_{VLA})$, also conditioned on VLA actions.
    \item $\boldsymbol{RL^2}$ (Ours): actions are generated by applying compositional steering to the base VLA samples (Sec.~\ref{sec:method}).
\end{enumerate}

Similar to existing test-time scaling studies~\cite{kwok2025robomonkey,kwok2026scaling}, we find that the relationship between the normalized action error and the number of generated action samples follows an exponential power law, assuming an oracle verifier.
For failure tuples, $RL^2$ compositional steering exhibits the strongest scaling behavior with significantly lower action error, indicating the importance of diversity during failure.
Even simple diversity-inducing methods like \textit{RBF} perform competitively, but fall behind $RL^2$ steering, likely because diversity is generated from the same base VLA. 
Most of the other methods outperform the \textit{Rephrase} baseline only by a tiny margin.

In contrast, under success tuples, most methods exhibit the opposite trend.
They consistently perform worse than the \textit{Rephrase} baseline, with RBF and $RL^2$ among the worst. 
These results indicate that diversity-inducing methods like compositional steering are most useful when the base VLA is likely to fail; RL-induced diversity can offer failure-avoidant or recovery actions beyond dominant VLA failure modes, but can unnecessarily perturb already-accurate actions when success is likely.
To the best of our knowledge, we are the first work to establish distinct test-time scaling laws comparing steering methods separately for success and failure tuples.

\section{Method}
\label{sec:method}

\begin{figure*}[!t]
\vspace{-0.1mm}
\centering
\includegraphics[width=1.0\textwidth]{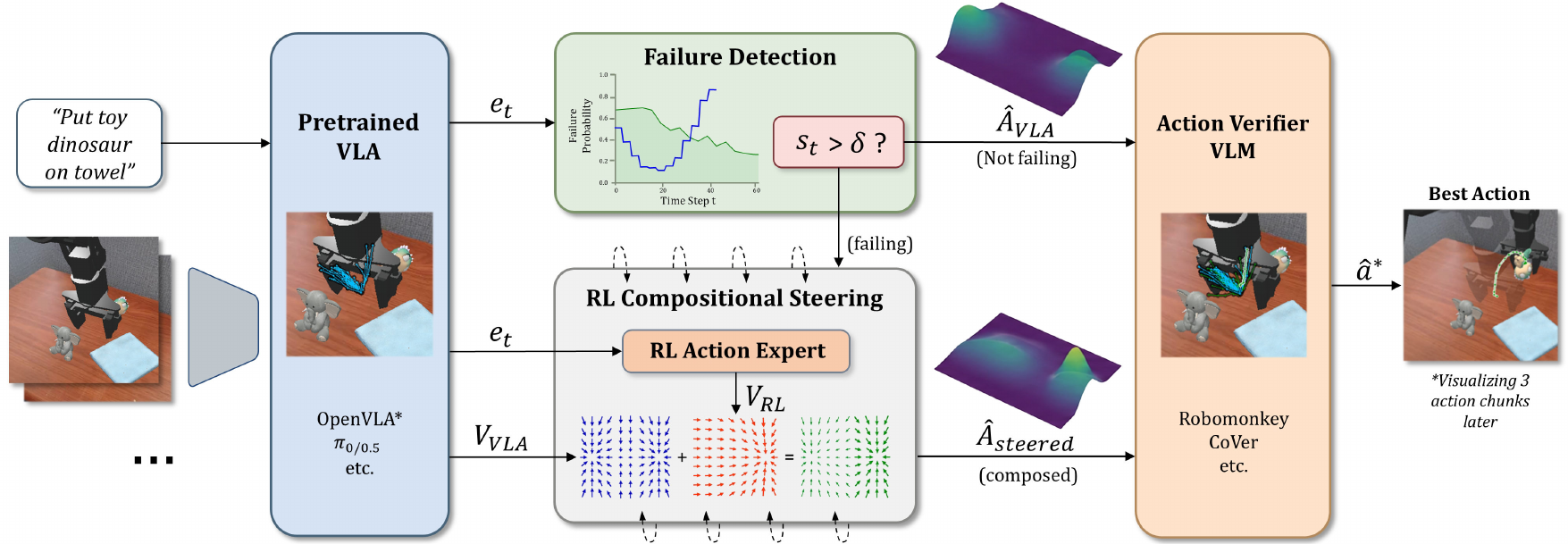}
\vspace{-7.5mm}
\caption{\textbf{$RL^2$ Framework (Flow-Matching VLAs):} \textbf{(i)} For flow-matching-based VLAs like $\pi_0$~\cite{black2026pi0}, given language instructions and observations, $RL^2$ first extracts action expert latents ($e_t$) from its frozen VLA action expert. 
\textbf{(ii)} The latents $e_t$ is fed into the RL flow-matching steering policy, which provides guidance velocities $V_{RL}$ that are composed with the VLA flow velocities $V_{VLA}$ at each flow-matching step to generate diverse and high-quality action candidates. 
\textbf{(iii)} In parallel, $e_t$ is passed into a failure detection module that determines whether steering should be enabled: if failure is detected, $RL^2$ composes the VLA and RL velocities; otherwise, it falls back to the base VLA action distribution. 
\textbf{(iv)} Finally, a verifier scores the generated candidates and selects the best action to execute
(*for autoregressive VLAs like OpenVLA~\cite{kim2024openvla}, we use Gaussian perturbation~\cite{kwok2025robomonkey} to perform RL compositional steering).
}
\vspace{-3.5mm}
\label{fig:method_overview}
\end{figure*}

We introduce $RL^2$, a framework to answer the question of \textbf{how} and \textbf{when} we should steer our pretrained VLAs towards diverse action candidates. 
More specifically, we enhance our action samples with RL compositional steering, only when our VLA is deemed by an external failure detection module to be failing.
We present the overall framework and pseudo-code of $RL^2$ in Fig.~\ref{fig:method_overview} and Algorithm~\ref{alg:rl2} respectively.

\subsection{RL Latent Policy Training}
To generate a differentiable guidance function $g(o,a,\ell)$ for steering VLA action samples, we first train a steering RL flow-matching policy $\pi_{\mathrm{RL}}(a_{t:t+H}\mid e_t)$ conditioned on latents extracted from the VLA action expert like in~\cite{gu2026safe}. 
We describe how these latents are aggregated into a single embedding vector $e$ in Sec.~\ref{sec:failure_detection_for_adaptive_steering}. 
To construct training data for RL policies conditioned on $e$, we augment existing large-scale datasets, namely BridgeV2~\cite{walke2023bridgedata} and DROID~\cite{khazatsky2024droid}, with policy latents to obtain $\mathcal{D}=\{o_i, a_i, \ell_i, e_i\}_{i=1}^{N}$.
Specifically, we iterate through each dataset tuple, pass its observations through the VLA, and extract the resulting embeddings. 

Training flow-matching actor-critic networks is challenging because direct backpropagation from the critic $Q(s, a)$ through the policy's multi-step flow-matching process is numerically unstable~\cite{park2025flow}. 
To mitigate this, we leverage QAM~\cite{li2026qam}, which replaces unstable backpropagation with a stable, step-wise matching objective.
QAM utilizes a lean adjoint state $\tilde{g}_t$ to act as a time-dependent guidance signal. This state is computed backwards from the terminal action $a_1$ via a reverse ODE using a fixed behavior prior $f_\beta$, an inverse temperature $\tau$, and with terminal condition $\tilde{g}_1=-\tau\nabla_{a_1}Q(s,a_1)$:
\begin{equation}
d\tilde{g}_t = -\nabla_{a_t}[2f_{\beta}(s,a_t,t)-a_t/t]\tilde{g}_t\,dt
\end{equation}

The policy's target velocity field $f_\theta$ is then optimized to align with this signal by minimizing the matching loss $L_{AM}(\theta)=$
\begin{equation}
\mathbb{E}_{s,\{a_t\}}\int_{0}^{1}\left\|\frac{2(f_{\theta}(s,a_t,t)-f_{\beta}(s,a_t,t))}{\sigma_t}+\sigma_t\tilde{g}_t\right\|^2_2\,dt,
\end{equation}
where $\sigma_t = \sqrt{2(1-t)/t}$ is a fixed noise schedule. This formulation ensures that the policy converges to the optimal behavior-regularized distribution $\pi(a|s) \propto \pi_\beta(a|s) \exp(\tau Q(s, a))$ while maintaining full expressivity through a stable, first-order optimization path.
In practice, and consistent with the findings in \cite{li2026qam}, we find that simply adjusting the inverse temperature $\tau$ is sufficient to achieve superior performance over behavior cloning (BC) baselines (see Sec.~\ref{sec:rl_vs_bc_vs_latents} for ablations).

Note that our $RL^2$ framework is compatible with different RL methods that are composable with the VLA.
This includes non-flow-based methods like the policy trained in V-GPS~\cite{nakamoto2024steering} via Conservative Q-Learning (CQL)~\cite{aviral2020cql} that are compatible with autoregressive VLAs such as OpenVLA~\cite{kim2024openvla}.
We train our RL policies on two NVIDIA L40s GPUs for 500k-1M steps, with training times largely determined by the VLA latents dimensionality (details in Appendix~\ref{app:steering_policy_details}).

\subsection{Compositional Steering}
In order to steer the frozen VLA action distribution $\pi_{\mathrm{VLA}}(a_{t:t+H}\mid o_t, \ell_t)$ toward candidates that are more diverse and of higher quality, we use the RL flow-matching head as an auxiliary steering function $\pi_{\mathrm{RL}}(a_{t:t+H}\mid e_t)$.
Following~\cite{cao2026compose}, we inject this guidance during each step of the VLA flow-matching process by composing the VLA and RL policies as a weighted average of their velocity fields. 
Specifically, if $v_{\mathrm{VLA}}$ denotes the VLA flow-matching velocity and $v_{\mathrm{RL}}$ denotes the RL-induced steering velocity, the enhanced VLA velocity is $v_{\mathrm{comp}} \leftarrow w v_{\mathrm{VLA}} + (1-w) v_{\mathrm{RL}}$. 
We draw composition weights $w$ from a Gaussian distribution ($\mu=\text{0.5}$, $\sigma=\text{0.25}$), such that action samples can benefit from both the demonstration-aligned action priors from VLAs and the diversity from RL steering by varying extents. 

For autoregressive VLAs such as OpenVLA~\cite{kim2024openvla}, they cannot be steered similarly using QAM because both action heads must be flow-matching policies. 
We instead adopt the Gaussian perturbation strategy from~\cite{kwok2025robomonkey}; we first fit a Gaussian distribution over an equal mix of VLA and RL action samples, and then sample a larger action batch from this fitted distribution.

\setlength{\textfloatsep}{6pt}
\begin{algorithm}[t]
\caption{$RL^2$: Adaptive RL Compositional Steering}
\label{alg:rl2}
\textit{Notation:} Uppercase symbols represent batches of size $N$ (e.g., $\hat{A}_t=\{\hat{a}_t^i\}_{i=1}^{N}$).
\begin{algorithmic}[1]
\REQUIRE Base VLA $\pi_{\mathrm{VLA}}$, RL steering policy $\pi_{\mathrm{RL}}$, failure detector $f_{\psi}$, CP threshold $\delta_t$, verifier $\mathcal{V}_{\theta}$, instruction $l_t$, observation $o_t$, batch size $N$
\STATE Extract VLA feature embedding $e_t$ {\scriptsize\textcolor{gray}{// {from action expert}}}
\STATE Predict LSTM failure score $s_t \leftarrow f_{\psi}(e_{0:t})$
\IF{$s_t > \delta_t$ {\scriptsize\textcolor{gray}{// failure}}}
    \STATE Sample compose weights $W \sim \mathcal{N}(0.5,0.25)$ {\scriptsize\textcolor{gray}{// {clipped [0,1]}}}
    \STATE Initialize noisy actions $\mathcal{T}_T \sim \mathcal{N}(0,I)$
    \FOR{$k = T,\ldots,1$ {\scriptsize\textcolor{gray}{// flow matching steps}}}
        \STATE $V_{\mathrm{VLA}} \leftarrow \pi_{\mathrm{VLA}}(\mathcal{T}_k,k,o_t,l_t)$
        \STATE $V_{\mathrm{RL}} \leftarrow \pi_{\mathrm{RL}}(\mathcal{T}_k,k,e_t)$ 
        \STATE $V_{\mathrm{comp}} \leftarrow WV_{\mathrm{VLA}} + (1-W)V_{\mathrm{RL}}$ \algcomment{velocity composition}
        \STATE $\mathcal{T}_{k-1} \leftarrow \mathrm{Step}(\mathcal{T}_k,V_{\mathrm{comp}},k)$ {\scriptsize\textcolor{gray}{// {flow step}}}
    \ENDFOR
    \STATE $\hat{A}_t \leftarrow \mathcal{T}_0$ {\scriptsize\textcolor{gray}{// {steered action samples}}}
\ELSE
    \STATE Sample VLA candidates $\hat{A}_t \sim \pi_{\mathrm{VLA}}(\cdot \mid o_t,l_t)$
\ENDIF
\STATE Score candidates $R_t \leftarrow \mathcal{V}_{\theta}(o_t,\hat{A}_t,l_t)$
\STATE Select $\hat{a}_t^{*} \leftarrow \hat{a}_t^{\,n^*}$, where $n^* = \arg\max_{n} R_t^n$ {\scriptsize\textcolor{gray}{// {best action}}}
\STATE Execute $\hat{a}_t^{*}$
\end{algorithmic}
\end{algorithm}

\subsection{Failure Detection for Adaptive Steering}
\label{sec:failure_detection_for_adaptive_steering}
\textbf{Multi-Task Failure Detector:}
As established in Sec.~\ref{sec:nrmse_inference_scaling_law}, compositional steering is most effective only when the base VLA is likely to fail, but can be unnecessary or even detrimental when the base policy already produces feasible actions. 
We therefore introduce a failure-detection trigger that enables steering only at timesteps when the VLA is detected to be failing, while otherwise falling back to the original VLA action distribution. 
We instantiate this trigger with SAFE~\cite{gu2026safe}, a lightweight multitask failure detector conditioned on informative internal VLA features.
For each task, we collect $100$ rollouts per seed across three random seeds, yielding trajectories that contain both successes and failures, and split them into training and validation sets $\mathcal{D}_{\mathrm{train}}$ and $\mathcal{D}_{\mathrm{val}}$ using a $60\%/40\%$ split. 
Following SAFE, we adopt an LSTM detector $f_{\psi}$ that takes in VLA features up to the current timestep and outputs a failure score $s_t=f_{\psi}(e_{0:t})\in[0,1]$. 
The per-timestep feature $e_t$ is obtained by extracting hidden-state vectors from the VLA action expert; we ablate different aggregation strategies (Appendix~\ref{app:vla_feature_extraction}) for converting these internal features into a single embedding vector $e_t$ and select the best one according to $\mathcal{D}_{\mathrm{val}}$ performance.
We train $f_{\psi}$ using a binary cross-entropy loss applied across all timesteps, $\mathcal{L}_{\mathrm{BCE}}=\sum_i\sum_t\left[y_i\log s_t+(1-y_i)\log(1-s_t)\right]$.

\textbf{Conformal Prediction (CP):}
Steering is triggered by a simple threshold rule: at timestep $t$, we intervene whenever the predicted failure score exceeds a time-varying threshold $\delta_t$. 
We calibrate $\delta_t$ using conformal prediction (CP), adopting the one-sided time-varying CP formulation following~\cite{xu2024uncertainty}.
Given a rollout-level score sequence $s_{1:T}$ and significance level $\alpha\in(0,1)$, the CP band is defined as $C_{\alpha}=\{[\mathrm{lower}_t,\mathrm{upper}_t]:t=1,\ldots,T\}$, with $\mathrm{lower}_t=-\infty$ and $\mathrm{upper}_t=\mu_t+h_t$, where $\mu_t$ and $h_t$ are the time-varying mean failure score and conformal bandwidth, respectively. 
We calibrate this band using successful rollouts from $\mathcal{D}_{\mathrm{val}}$, so that any new successful rollout satisfies $s_t < \mu_t+h_t$ for all timesteps with probability at least $1-\alpha$. 
Note that while we train a single failure detector on all tasks, we calibrate per-task CP bands and a combined CP band for in-domain and OOD tasks respectively.
More details and illustrations can be found in Appendix~\ref{app:intervention_comparison}.

\begin{figure*}[!t]
\centering
\begin{minipage}[t]{0.48\textwidth}
\vspace{0pt}
\centering
\includegraphics[width=\linewidth]{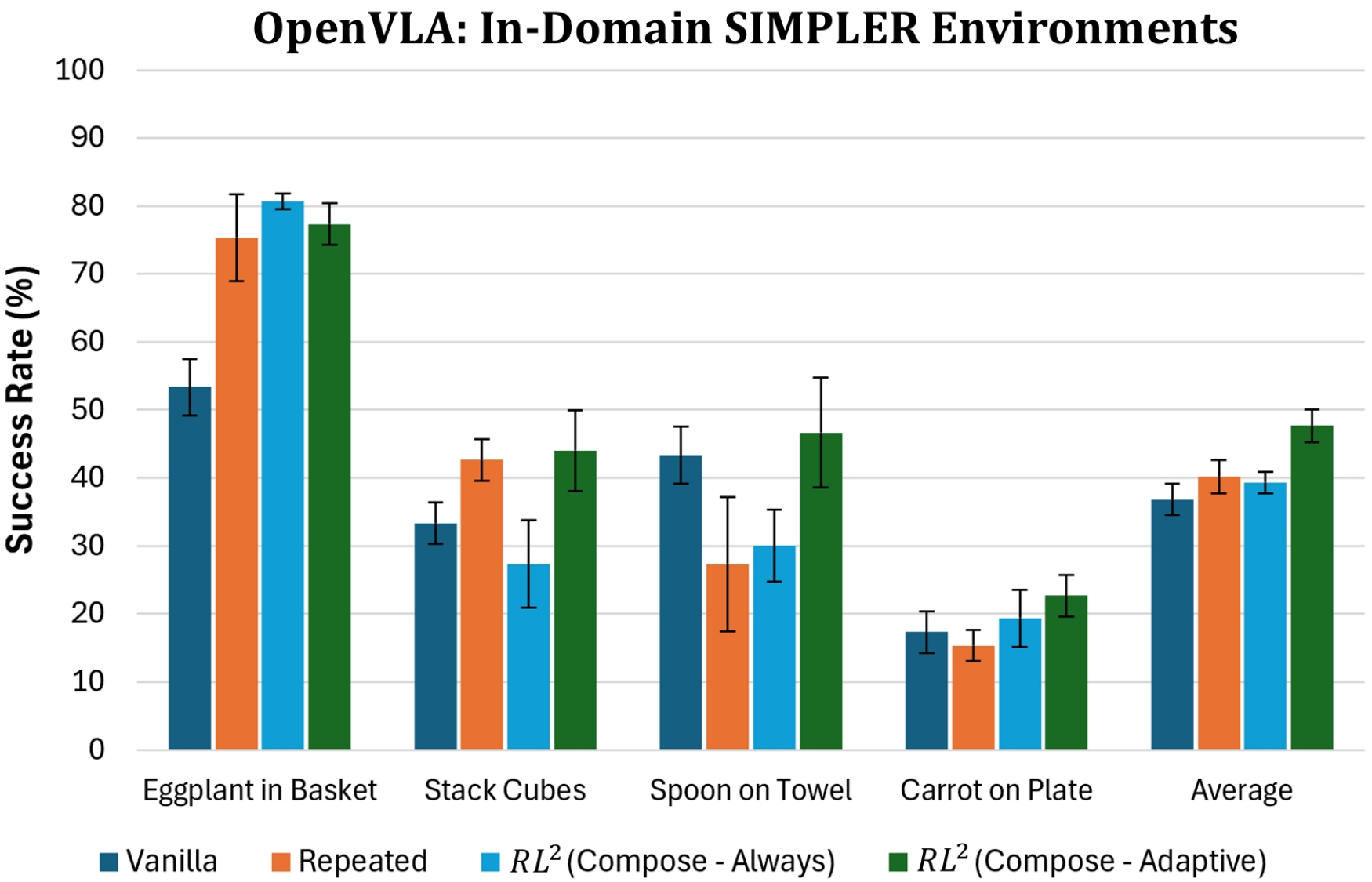}
\vspace{-6mm}
\captionof{figure}
{\textbf{SIMPLER In-Domain Evaluation:} Using OpenVLA~\cite{kim2024openvla}, adaptive $RL^2$ improves performance on standard SIMPLER tasks~\cite{li24simpler}, by up to +19.4\% in task-wise success rate (average +7.5\%) over \textit{Repeated} baseline.
}
\label{fig:openvla_simpler_iid_plot}
\end{minipage}
\hfill
\begin{minipage}[t]{0.48\textwidth}
\vspace{0pt}
\centering
\includegraphics[width=\linewidth]{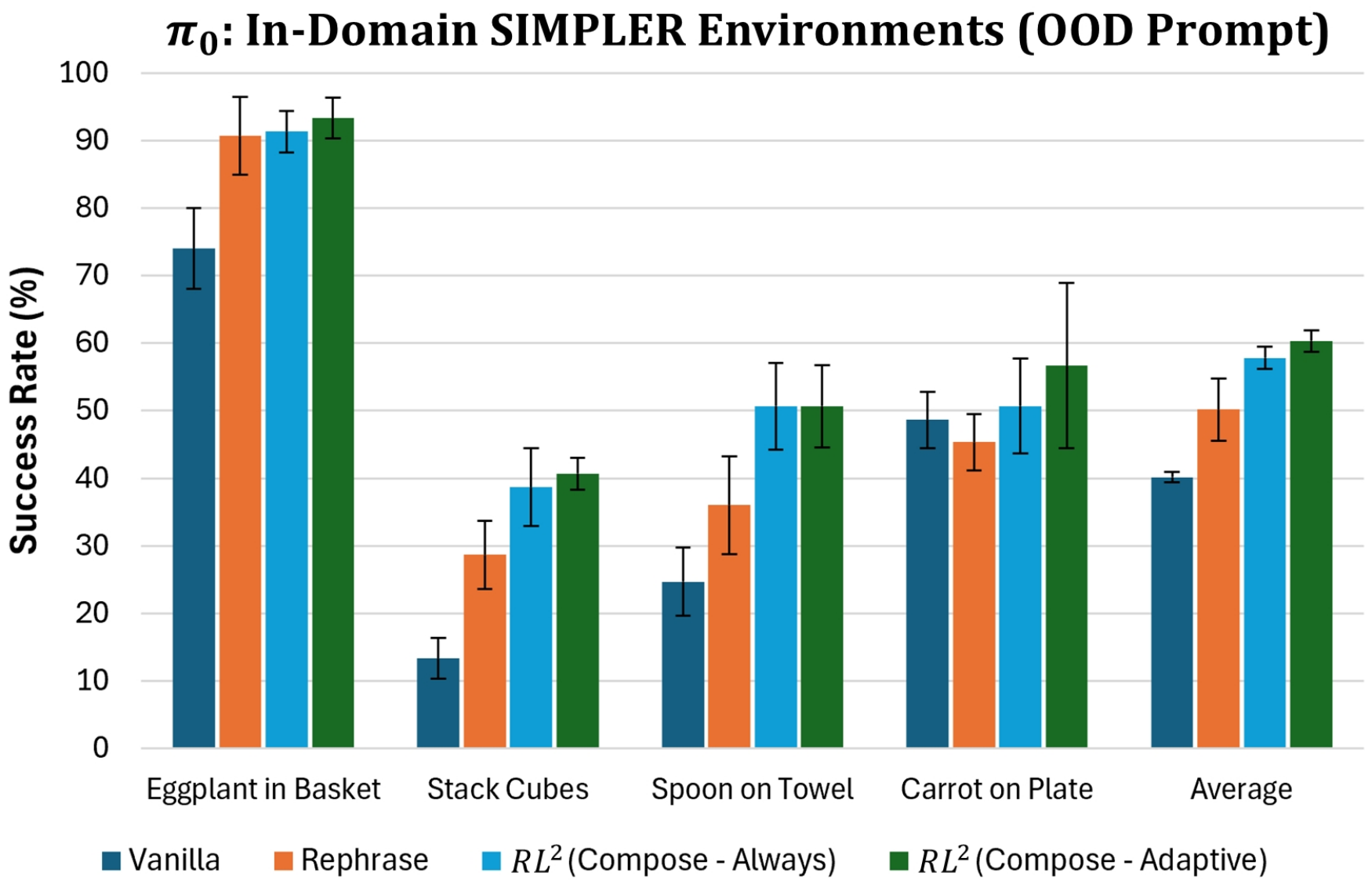}
\vspace{-6mm}
\captionof{figure}
{\textbf{SIMPLER OOD Prompt Evaluation:} Using $\pi_0$~\cite{black2026pi0}, adaptive $RL^2$ improves robustness under OOD language instructions~\cite{fang2025intact}, achieving up to +14.7\% in task-wise success rate (average +10.1\%) over \textit{Rephrase} baseline.}
\label{fig:pi0_simpler_rephrases_plot}
\end{minipage}
\vspace{-4mm}
\end{figure*}

\textbf{CP Alpha Selection Heuristic:}
In practice, we empirically observe that the best CP significance level $\alpha$ varies across tasks because $\alpha$ directly controls the frequency with which steering is triggered. 
Rather than exhaustively sweeping all possible $\alpha$ values for every task which would require costly evaluations (like in~\cite{cao2026compose} for composition weights), we instead introduce a \textit{test-time alpha selection heuristic}.
On $\mathcal{D}_{\mathrm{val}}$, we evaluate candidate $\alpha$ values by balanced accuracy, $\mathrm{BalAcc}(\alpha)=\frac{1}{2}(\mathrm{TPR}(\alpha)+\mathrm{TNR}(\alpha))$, where $\mathrm{TPR}$ measures the fraction of failed rollouts correctly flagged as failures and $\mathrm{TNR}$ measures the fraction of successful rollouts correctly flagged as success. 
We then perform a calibration sweep for the top three $\alpha$ values under this metric and select the best-performing value for final evaluation.  
We find that our choice of balanced accuracy as an $\alpha$ selection metric correlates well with downstream task performance (Sec.~\ref{sec:alpha_selection_heuristics_results}).
More details and illustrations of our heuristic can be found in Appendix~\ref{app:alpha_selection_heuristics}.

\subsection{Action Verification}
Given a set of $N$ candidate actions $\hat{A} = \{ \hat{a}_t^{\,1}, \ldots, \hat{a}_t^{\,N}\}$, where $\hat{a}$ refers to either VLA or steered action samples, an action verifier serves as a reward model to select the best action sample to execute for each timestep $t$.
More specifically, the verifier assigns each candidate a score $r_t^n=\mathcal{V}_{\theta}(o_t,\hat{a}_t^{\,n}, l_t)$, and selects the highest scoring sample $\hat{a}_t^* = \arg\max_{n} (r_t^n)$.
$RL^2$ is designed to be adaptable to different types of action verifiers. This ranges from verifiers trained via preference learning such as RoboMonkey~\cite{kwok2025robomonkey} to verifiers trained via contrastive learning such as CoVer~\cite{kwok2026scaling}.

\section{Experiments}
\label{sec:experiments}
The primary objective of our experiments is to test the robustness and generalization of $RL^2$ to challenging or OOD tasks,
while providing a modular interface compatible with various pretrained VLAs and verifiers. 
While this work tackles several research questions, the main ones are:
\begin{itemize}
    \item Does compositional steering improve task performance?
    \item How does using latents and RL enable effective steering?
    \item How well does $RL^2$ scale with different numbers and types of action samples?
\end{itemize}

\subsection{Experimental Setup}
We evaluate $RL^2$ across a diverse set of simulation benchmarks, VLAs, and verifiers. 
In addition to testing the robustness of our framework on standard benchmarks, we also focus on tasks that assess $RL^2$'s generalization to OOD language instructions and task environments. 
To achieve this, we adopt the experimental framework from prior works on test-time scaling~\cite{kwok2025robomonkey,kwok2026scaling}.
These tasks are visualized in Fig.~\ref{fig:iid_vs_ood_expt_examples}, and more details can be found in Appendix~\ref{app:additional_task_description}.

\textbf{SIMPLER (Original):}
The SIMPLER benchmark~\cite{li24simpler} is a simulation environment with in-domain manipulation tasks mimicking demonstration setups from large-scale datasets.
In this environment, we test OpenVLA~\cite{kim2024openvla} on four of their standard in-domain tasks from BridgeV2~\cite{walke2023bridgedata} (WidowX robot) with their original task prompts. 
We use the policy trained in V-GPS~\cite{nakamoto2024steering} via CQL~\cite{aviral2020cql} as our RL steering policy and RoboMonkey~\cite{kwok2025robomonkey} as our verifier.

\textbf{SIMPLER (OOD):}
We test $\pi_0$ using the SIMPLER benchmark with the same four in-domain tasks, but with challenging red-teaming instructions~\cite{Karnik2024EmbodiedRT} to simulate OOD base language prompts, via the implementation in~\cite{fang2025intact}.
We also introduce four additional OOD task environments from Interleave-VLA~\cite{fan2026interleavevla}, which offer a mix of varying backgrounds, task objects, and the presence of distractors.
All of these OOD tasks assess if VLAs can enhance task performance using VLM-generated rephrases. 
Here, we use QAM~\cite{li2026qam} as our RL flow-based steering policy and CoVer~\cite{kwok2026scaling} as our verifier.

\textbf{PolaRiS (OOD):}
We test $\pi_{0.5}$ on the PolaRiS benchmark~\cite{jain2025polaris}, which offers high-fidelity Gaussian-splat simulation environments with a Franka Emika Panda robot.
We evaluate our approach on three tasks adapted from the DROID dataset~\cite{khazatsky2024droid}, but with OOD red-teaming base prompts.
These tasks contain many other distractor objects in the scene, and assess if VLAs can enhance task performance using VLM-generated rephrases.
Likewise, we use QAM~\cite{li2026qam} as our RL flow-based steering policy and CoVer~\cite{kwok2026scaling} as our verifier.

\textbf{Baselines:}
To evaluate $RL^2$, we compare our approach with several baselines: (1) \textbf{Vanilla} represents either vanilla OpenVLA or $\pi_{0/0.5}$, (2) \textbf{Repeated/Rephrase} samples multiple actions from the VLA via repeated sampling~\cite{kwok2025robomonkey} or VLM-generated rephrases~\cite{kwok2026scaling}, (3) \textbf{$RL^2~\text{(Compose-Always)}$} uses compositional steering on multiple actions sampled from the VLA, and (4) \textbf{$RL^2~\text{(Compose-Adaptive)}$} uses compositional steering only when the VLA is flagged as failing by SAFE~\cite{gu2026safe} but uses \textit{Repeated}/\textit{Rephrase} otherwise. 
Note that for OpenVLA, we follow~\cite{kwok2025robomonkey} which employs Gaussian perturbation by fitting a Gaussian distribution over 9 action samples (VLA and/or composed), then re-sampling 32 actions from it.
In addition, for $\pi_{0/0.5}$, we follow~\cite{kwok2026scaling} which uses 8 rephrases with 5 action samples each. 
We run all experiments 50 times each and report the mean and standard deviation across 3 random seeds.
We will release our code with the final paper.

\begin{figure}[!t]
\centering
\includegraphics[width=\columnwidth]{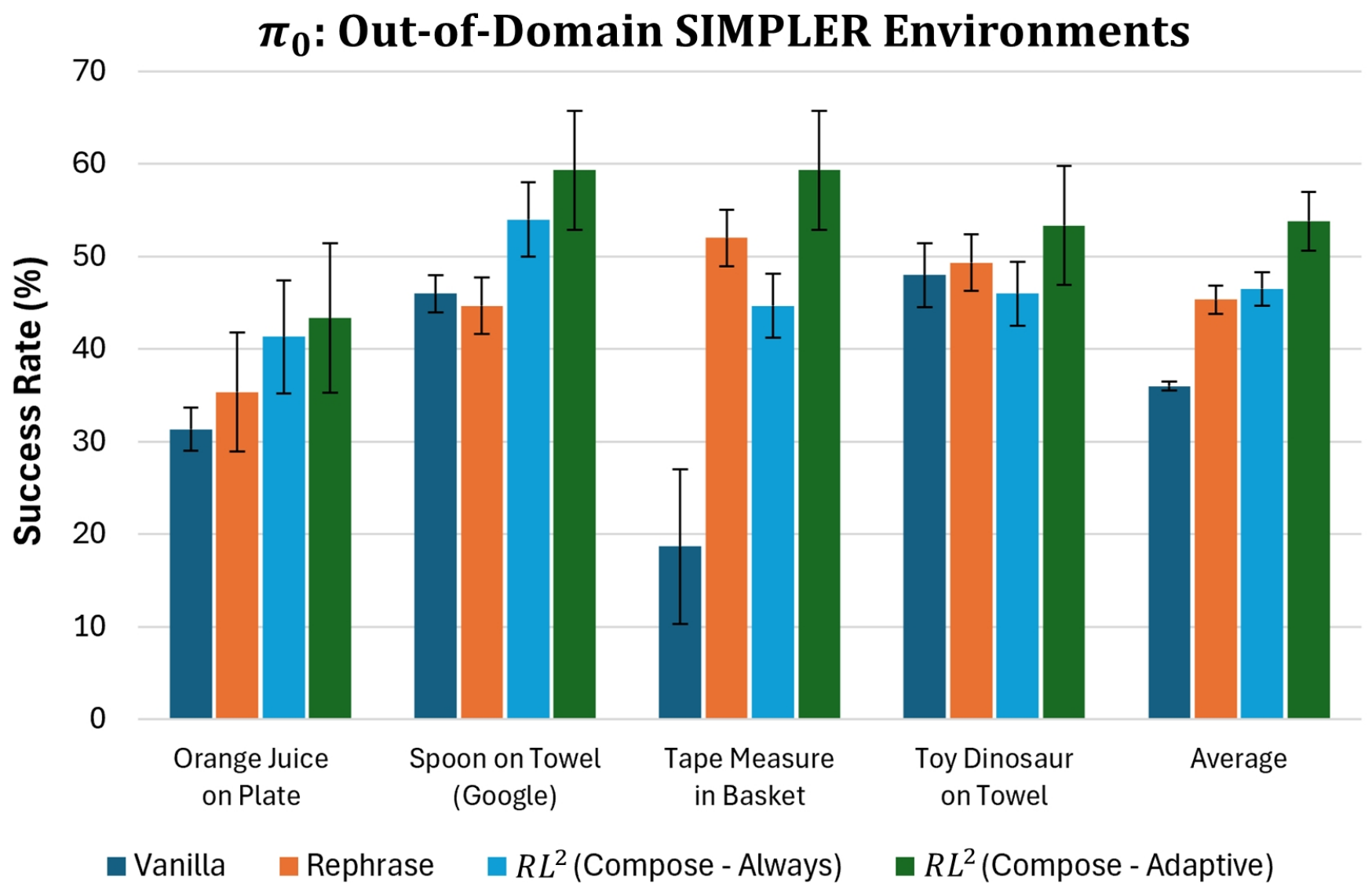}
\vspace{-6mm}
\caption{
\textbf{SIMPLER OOD Environment Evaluation:} Using $\pi_0$~\cite{black2026pi0}, adaptive $RL^2$ improves robustness in OOD environments~\cite{fang2025intact}, by up to +14.6\% in task-wise success rate (average +8.5\%) over the strongest \textit{Rephrase} baseline.
}
\label{fig:pizero_sim_ood_plot}
% \vspace{-4mm}
\end{figure}

\subsection{Simulation Results}

We summarize our simulation results for both in-domain and OOD tasks, and report average and task-wise performance gains of up to +10.9\% and +19.4\% respectively over our strongest baseline. 
We believe these gains are significant given that we are only using offline RL.
More baseline comparisons can be found in our NRMSE analysis (Sec.~\ref{sec:nrmse_inference_scaling_law}).

\vspace{0.2em}
\subsubsection{\textbf{Compositional steering is robust to standard and OOD language prompts}}
\label{sec:sim_expts_rephrases}
To illustrate the robustness of $RL^2$ towards standard and OOD base prompts on tasks seen by the VLA, we present our simulation results for OpenVLA, $\pi_0$, and $\pi_{0.5}$ in Figs.~\ref{fig:openvla_simpler_iid_plot} and~\ref{fig:pi0_simpler_rephrases_plot}, and Table~\ref{tab:pi05_polaris_main_results}. 
Based on the average performance across all benchmarks, we observe a general improvement in success rates with our adaptive $RL^2$ framework. 
For OpenVLA, adaptive $RL^2$ achieves an average performance gain of +7.5\% (task-wise +19.4\%) over the strongest \textit{Repeated} baseline across SIMPLER in-domain tasks.
For $\pi_0$, adaptive $RL^2$ achieves an average performance gain of +10.1\% over the strongest \textit{Rephrase} baseline across SIMPLER tasks under OOD language prompts, with +14.7\% gain for the highest performing task (\textit{Spoon on Towel}).
For $\pi_{0.5}$, adaptive $RL^2$ achieves an average performance gain of +10.9\% and +7.8\% on success and progress rates respectively, over the strongest \textit{Rephrase} baseline across PolaRiS tasks under OOD language prompts (+17.3\% success rate for \textit{Move Latte Cup} task).

These results indicate the robustness of $RL^2$ to in-domain and OOD language prompts.
We attribute this to the diversity introduced by RL compositional steering during failure states.
As seen in Fig.~\ref{fig:rl2_viz_cover_samples}, where we visualize action chunks from $\pi_0$ using rephrases for the \textit{Spoon on Towel} task, the action samples tend to be uniformly directed away from the spoon. 
In contrast, the steered action samples tend to be more diverse and offer some higher-quality action samples towards the spoon, one of which is selected by the CoVer verifier.  
Across the episode, we observe that $\pi_0$ with rephrases leads to oscillations above the spoon, while $RL^2$ leads to accurate and decisive grasping of the spoon.

\begin{table}[!t]
\centering
\caption{
\textbf{PolaRiS OOD Prompt Evaluation:} Using $\pi_{0.5}$~\cite{intelligence2025pi05}, adaptive $RL^2$ improves robustness under OOD language instructions~\cite{fang2025intact}, achieving up to +17.3\% in task-wise success rate ($S$) and +12.4\% in task-wise progress rate ($P$) over \textit{Rephrase} baseline.
}
\vspace{-1mm}
\label{tab:pi05_polaris_main_results}
\scriptsize
\setlength{\tabcolsep}{3pt}
\begin{tabular*}{0.98\columnwidth}{@{\extracolsep{\fill}}llcccc@{}}
\toprule
\makecell[c]{\textbf{Task}} & \makecell[c]{\textbf{Metric}}
& \makecell[c]{\textbf{$\pi_{0.5}$}\\\textbf{(Vanilla)}}
& \makecell[c]{\textbf{$\pi_{0.5}$}\\\textbf{(Rephrase)}}
& \makecell[c]{\textbf{$RL^2$}\\\textbf{(Compose)}}
& \makecell[c]{\textbf{$RL^2$}\\\textbf{(Adaptive)}} \\
\midrule
\multirow{2}{*}{\makecell[l]{Move\\Latte Cup}} & $S(\%)$ & 18.7 & 48.7 & 55.3 & \textbf{66.0} \\
             & $P(\%)$ & 39.5 & 65.6 & 72.0 & \textbf{78.0} \\
\midrule
\multirow{2}{*}{\makecell[l]{Tape into\\Container}} & $S(\%)$ & 12.7 & 22.0 & 22.0 & \textbf{28.7} \\
          & $P(\%)$ & 34.6 & 41.3 & 40.9 & \textbf{46.9} \\
\midrule
\multirow{2}{*}{\makecell[l]{Pan\\Cleaning}} & $S(\%)$ & 11.5 & 24.7 & 24.0 & \textbf{33.3} \\
            & $P(\%)$ & 45.7 & 58.7 & 54.4 & \textbf{64.0} \\
\midrule
\multirow{2}{*}{Average} & $S(\%)$ & 14.3 (\(\pm 2.0\)) & 31.8 (\(\pm 3.4\)) & 33.8 (\(\pm 1.7\)) & \textbf{42.7 (\(\pm 2.3\))} \\
         & $P(\%)$ & 40.0 (\(\pm 0.7\)) & 55.2 (\(\pm 1.4\)) & 55.8 (\(\pm 1.2\)) & \textbf{63.0 (\(\pm 3.4\))} \\
\bottomrule
\end{tabular*}
\end{table}

\vspace{0.2em}
\subsubsection{\textbf{Compositional steering generalizes well to OOD task environments}}
\label{sec:sim_expts_ood}
To assess the generalization of $RL^2$ on task environments that were not seen by the VLA during training, we introduce four additional tasks within the SIMPLER environment with varying backgrounds, task objects, and
the presence of distractors (Fig.~\ref{fig:iid_vs_ood_expt_examples} and Appendix~\ref{app:additional_task_description}).
We benchmark using $\pi_0$ across these SIMPLER tasks under OOD language prompts, and observe an average performance gain of +8.5\% over the strongest $Rephrase$ baseline, with up to +14.6\% gain for the highest performing task (\textit{Spoon on Towel - Google}).
We can gain some insights into the strong performance of $RL^2$ in Fig.~\ref{fig:rl2_viz_cover_samples}, where we can observe that \textit{Rephrase} $\pi_0$ action samples tend to be directed mostly towards the wrong distractor object (i.e., toy elephant) during a failure state.
In contrast, the RL-steered action samples offer diversity that skews the overall action distribution more towards the main task object (i.e., dinosaur), leading to accurate and decisive grasping.

\begin{table*}[!t]
\centering
\begin{minipage}[t]{0.48\textwidth}
\vspace{0pt}
\centering
\caption{\textbf{Ablation studies:} RL training and VLA latents enhance compositional steering (*in-domain prompt for OpenVLA).}
\vspace{-2mm}
\label{tab:latent_rl_ablation}
\footnotesize
\begin{tabular}{lcccc}
\toprule
\multicolumn{1}{l}{\makecell[l]{\textbf{Model}}} & \makecell{\textbf{Latents}} & \makecell{\textbf{Method}} & \makecell{\textbf{OOD}\\\textbf{Prompts (\%)}} & \makecell{\textbf{OOD}\\\textbf{Tasks (\%)}} \\
\midrule
{$\pi_0$~\cite{black2026pi0}} & \cmark & RL & \textbf{57.8} (\(\pm 1.6\)) & \textbf{46.5} (\(\pm 1.8\)) \\
{(SIMPLER)} & \cmark & BC & 55.8 (\(\pm 3.0\)) & 42.0 (\(\pm 4.0\)) \\
\midrule
{$\pi_{0.5}$~\cite{intelligence2025pi05}} & \cmark & RL & \textbf{33.8} (\(\pm 1.7\)) & -- \\
{(PolaRiS)} & \cmark & BC & 29.7 (\(\pm 4.7\)) & -- \\
\midrule
OpenVLA*~\cite{kim2024openvla} & \cmark & RL & \textbf{39.3} (\(\pm 1.6\)) & -- \\
(SIMPLER) & \xmark & RL & 0.5 (\(\pm 0.5\)) & -- \\
\bottomrule
\end{tabular}
\end{minipage}
\hfill
\begin{minipage}[t]{0.48\textwidth}
\vspace{0pt}
\centering
\caption{\textbf{Ablation studies:} SAFE~\cite{gu2026safe} failure detector is an effective and timely trigger for compositional steering.}
\vspace{-2mm}
\label{tab:trigger_ablation}
\setlength{\tabcolsep}{8pt}
\renewcommand{\arraystretch}{1.09}
\begin{tabular}{lccc}
\toprule
\textbf{Model} & \textbf{Trigger} & \makecell{\textbf{OOD}\\\textbf{Prompts (\%)}} & \makecell{\textbf{OOD}\\\textbf{Tasks (\%)}} \\
\midrule
\multirow{3}{*}{
\shortstack[l]{$\pi_0$~\cite{black2026pi0}\\
(SIMPLER)}}
    & SAFE~\cite{gu2026safe}
    & \textbf{60.3} (\(\pm1.6\))
    & \textbf{53.8} (\(\pm3.2\)) \\
    & CoVer~\cite{kwok2026scaling}
    & 57.8 (\(\pm1.3\))
    & 50.3 (\(\pm1.2\)) \\
    & Always
    & 57.8 (\(\pm1.6\))
    & 46.5 (\(\pm1.8\)) \\
\midrule
\multirow{3}{*}{
\shortstack[l]{$\pi_{0.5}$~\cite{intelligence2025pi05}\\
(PolaRiS)}}
    & SAFE~\cite{gu2026safe}
    & \textbf{42.7} (\(\pm2.3\))
    & -- \\
    & CoVer~\cite{kwok2026scaling}
    & 39.8 (\(\pm1.8\))
    & -- \\
    & Always
    & 33.8 (\(\pm1.7\))
    & -- \\
\bottomrule
\end{tabular}
\end{minipage}%
\end{table*}

\begin{figure*}[t]
\centering
\begin{minipage}[t]{0.315\textwidth}
\vspace{0pt}
\centering
\includegraphics[width=\linewidth]{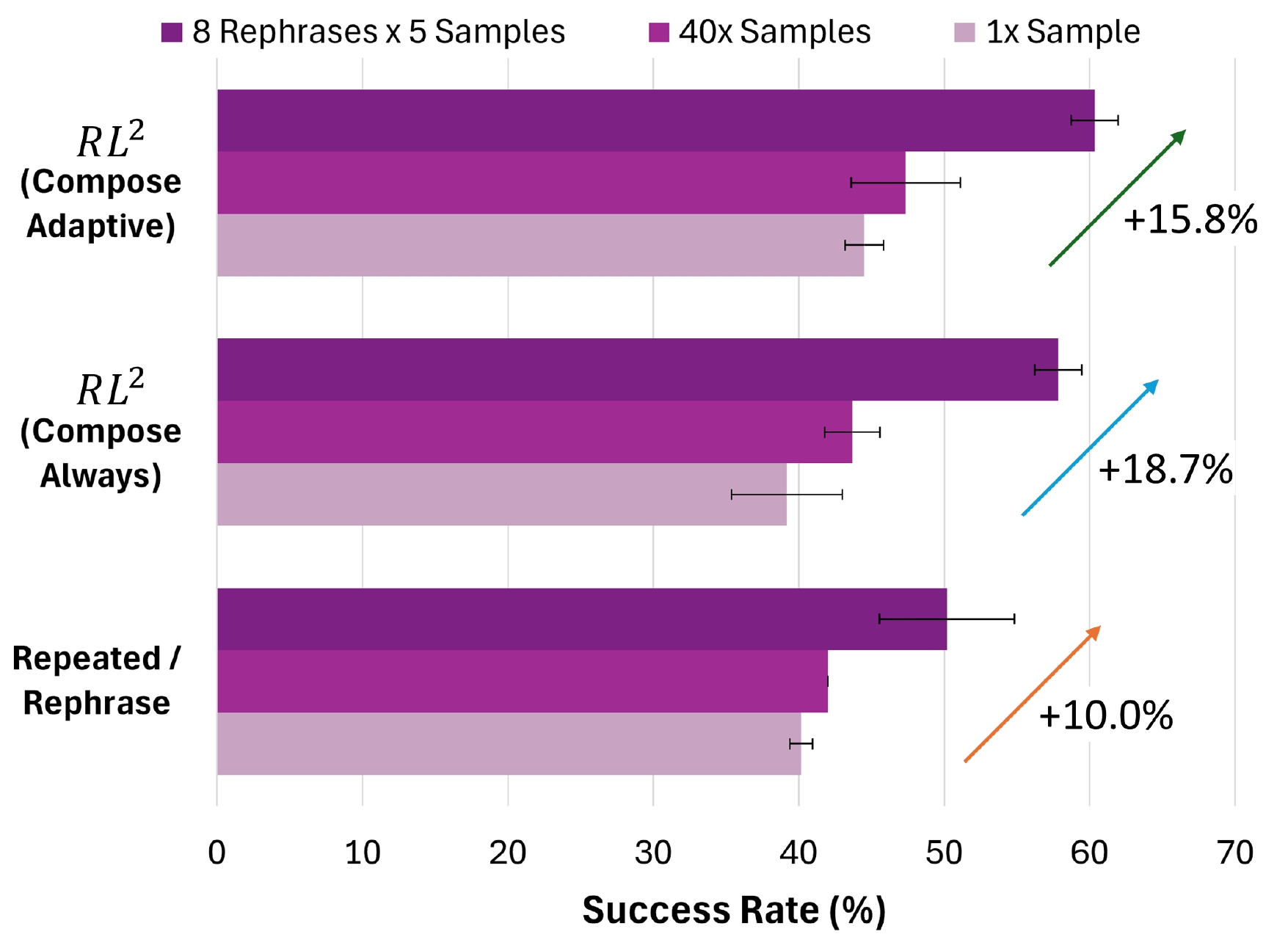}
\vspace{-8.3mm}
\captionof{figure}{\textbf{Scaling Samples and Rephrases: } 
$RL^2$ attains larger performance gain compared to baselines when scaling both samples and rephrases.
}
\label{fig:scaling_rephrases_and_samples}
\end{minipage}
\hspace{0.02\textwidth}%
\begin{minipage}[t]{0.315\textwidth}
\vspace{0pt}
\centering
\includegraphics[width=\linewidth]{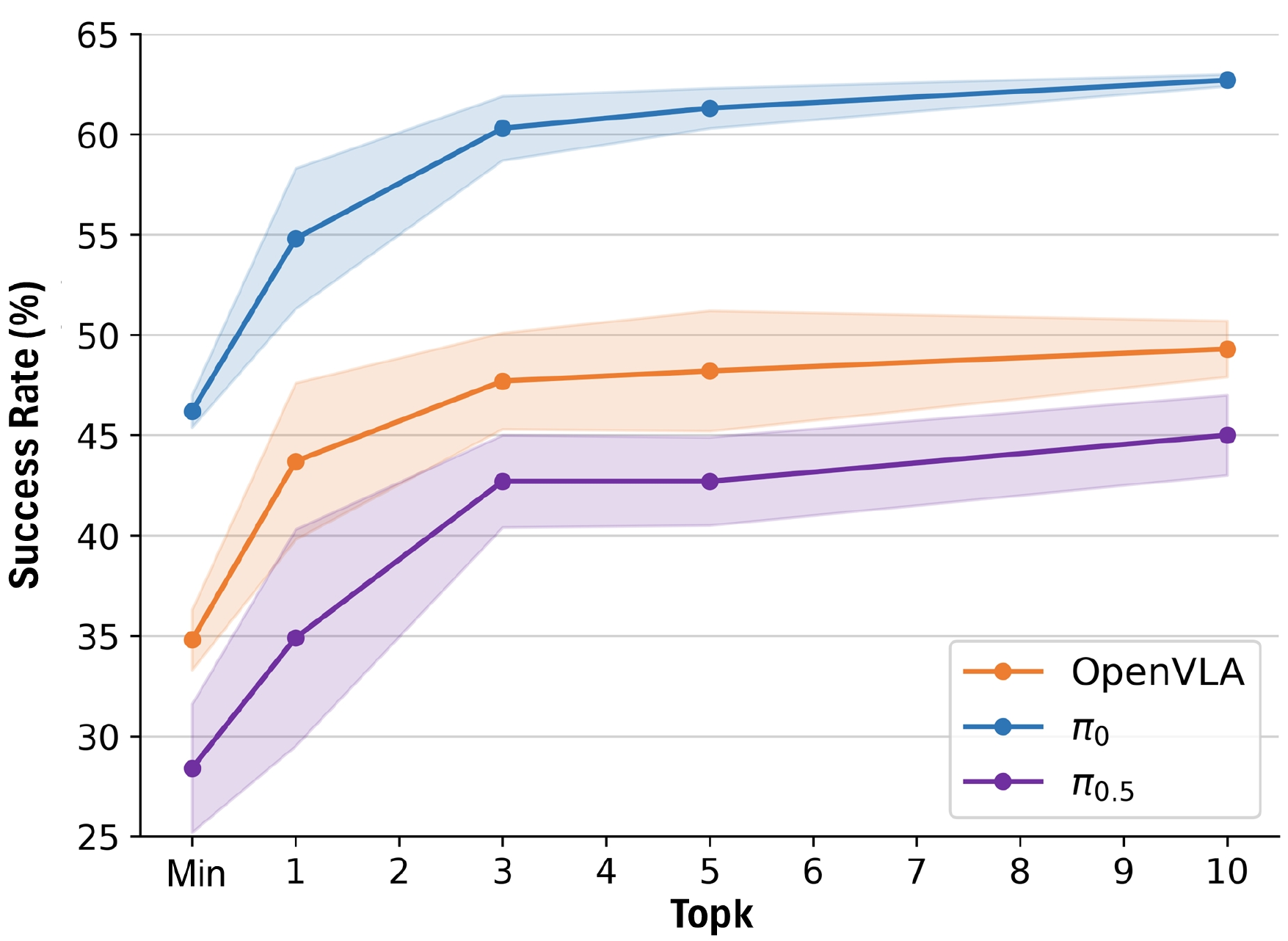}
\vspace{-8mm}
\captionof{figure}{\textbf{CP Alpha Selection Heuristic:} Our heuristic selects one of the top alphas within top-3 selection across all baselines.}
\label{fig:alpha_selection_heuristics}
\end{minipage}%
\hspace{0.025\textwidth}%
\begin{minipage}[t]{0.315\textwidth}
\vspace{0pt}
\centering
\scriptsize
\captionof{table}{\textbf{Scaling time analysis: } RL steering and failure detection remain efficient with increasing batch sizes\\(in ms, on NVIDIA RTX5090 GPU).}
\vspace{-2mm}
\setlength{\tabcolsep}{6pt}
\begin{tabular}{c|cccc}
\toprule
\makecell{\textbf{Batch}\\\textbf{Size}} & \makecell{$\pi_0$\\\cite{black2026pi0}} & \makecell{CoVer\\\cite{kwok2026scaling}} & \makecell{QAM\\\cite{li2026qam}} & \makecell{SAFE\\\cite{gu2026safe}} \\
\midrule
1   & 232  & 96  & 12 & 1 \\
2   & 243  & 100 & 15 & 1 \\
4   & 265  & 102 & 16 & 1 \\
8   & 320  & 107 & 17 & 2 \\
16  & 434  & 111 & 21 & 2 \\
32  & 724  & 114 & 27 & 2 \\
64  & 1541 & 116 & 40 & 2 \\
128 & 3093 & 145 & 48 & 2 \\
\bottomrule
\end{tabular}
\label{tab:inference_time_analysis}
\end{minipage}%
\vspace{-4mm}
\end{figure*}

\subsection{Ablations}
We perform a comprehensive series of ablation studies to verify that all components of $RL^2$ are crucial in improving downstream performance across various benchmarks.

\vspace{0.2em}
\subsubsection{\textbf{Adaptiveness is crucial for effective steering}}
To assess the importance of adaptiveness in compositional steering, we perform an ablation study by removing the failure detection module in $RL^2$ (Figs.~\ref{fig:openvla_simpler_iid_plot} and~\ref{fig:pi0_simpler_rephrases_plot}, Table~\ref{tab:pi05_polaris_main_results}).
While non-adaptive steering performs nearly on-par with adaptive steering for $\pi_0$, the broader trend across OpenVLA and $\pi_{0.5}$ indicates that adaptive steering yields significantly stronger gains.
More specifically, we observe an average and task-wise performance gain of up to +8.9\% and +16.7\% respectively.
This aligns well with our NRMSE analysis (Sec.~\ref{sec:nrmse_inference_scaling_law}) where compositional steering is most beneficial only when applied to failure states.

\vspace{0.2em}
\subsubsection{\textbf{Using latents and RL results in effective steering}}
\label{sec:rl_vs_bc_vs_latents}
We first perform an ablation by modifying the training recipe from RL to Behavior Cloning (BC). 
We measure the performance of composition at all times for $\pi_{0/0.5}$ in Table~\ref{tab:latent_rl_ablation}, which indicates that using RL consistently outperforms BC by up to +4.5\%.
This supports our intuition that a different yet effective training recipe like RL can introduce the diversity required for better verifier action selection during VLA failure.

In addition, we perform an ablation for $\pi_{0/0.5}$ by conditioning our steering policy on raw observations instead of VLA latents (Table~\ref{tab:latent_rl_ablation}).
However, we were unable to successfully train QAM on raw observations in BridgeV2 as it was originally optimized for proprioceptive inputs on OGBench~\cite{park2025ogbench}.
Hence, we test this with OpenVLA instead, using the CQL policy trained in V-GPS as its RL steering policy that is optimized for vision-language inputs.
We observe that steering via latents achieves a performance gain of +38.8\% over steering without latents.
Notably, the vanilla V-GPS formulation conditioned on raw observations performed poorly at 0.5\%.
This large gap is likely due to the use of latents from the VLA action expert, which is significantly more expressive~\cite{gu2026safe} than features from small encoders like ResNet used by vanilla V-GPS.

\vspace{0.2em}
\subsubsection{\textbf{Failure detection methods outperform other triggers for adaptive compositional steering}}

To assess the importance of SAFE as an effective trigger for compositional steering, we perform an ablation study by using an external VLM as the trigger instead.
More specifically, we use the CoVer verifier~\cite{kwok2026scaling} as our VLM for both failure detection and action selection.
Similar to our $RL^2$ setup in Sec.~\ref{sec:failure_detection_for_adaptive_steering}, we collect online rollouts with CoVer verifier scores in order to generate CP bands for failure detection. 
As seen from Table~\ref{tab:trigger_ablation}, using CoVer as a failure detector generally leads to improvements over the baseline of steering at all times. 
Nevertheless, using SAFE still performs up to +3.5\% better compared to using CoVer, likely because SAFE is conditioned on informative latents extracted directly from the action expert.  
Conversely, CoVer is a poorer failure detector, as seen from Appendix~\ref{app:intervention_comparison} where success and failure scores are too correlated and thus not as separable as compared to using SAFE.

\begin{figure*}[!t]
\centering
\begin{minipage}[t]{0.535\textwidth}
\vspace{0pt}
\centering
\includegraphics[width=\linewidth]{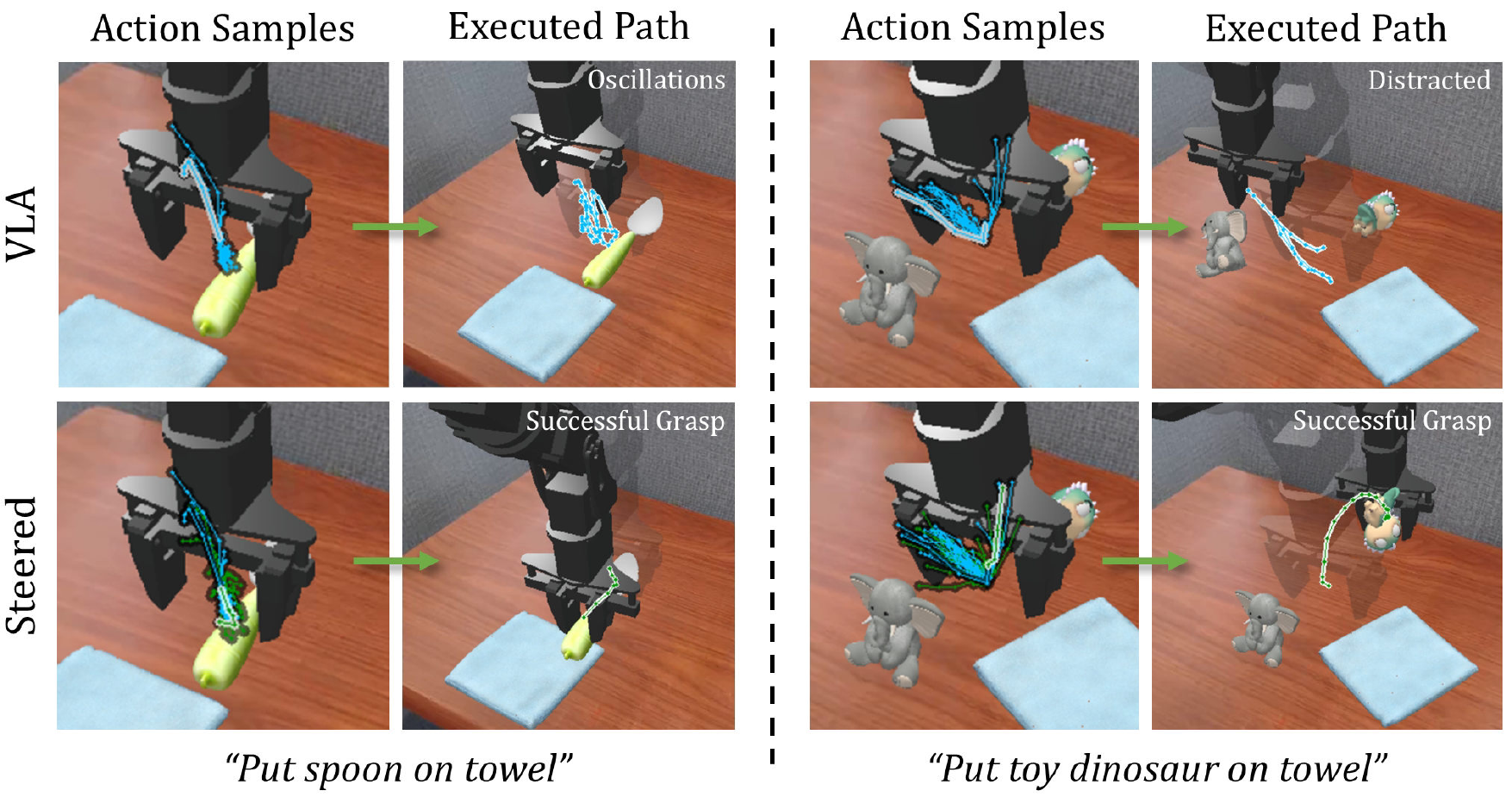}
\vspace{-7.5mm}
\captionof{figure}{\textbf{Action Samples Visualization ($\pi_0$):} During failure, VLA actions (blue) using rephrases tend to either be directed away from the task object or toward the distractor object, resulting in oscillatory behaviors. $RL^2$ performs compositional steering (green) to direct VLA samples toward the task object for successful grasping.}
\label{fig:rl2_viz_cover_samples}
\end{minipage}
\hspace{0.02\textwidth}
\begin{minipage}[t]{0.43\textwidth}
\vspace{0pt}
\centering
\includegraphics[width=\linewidth]{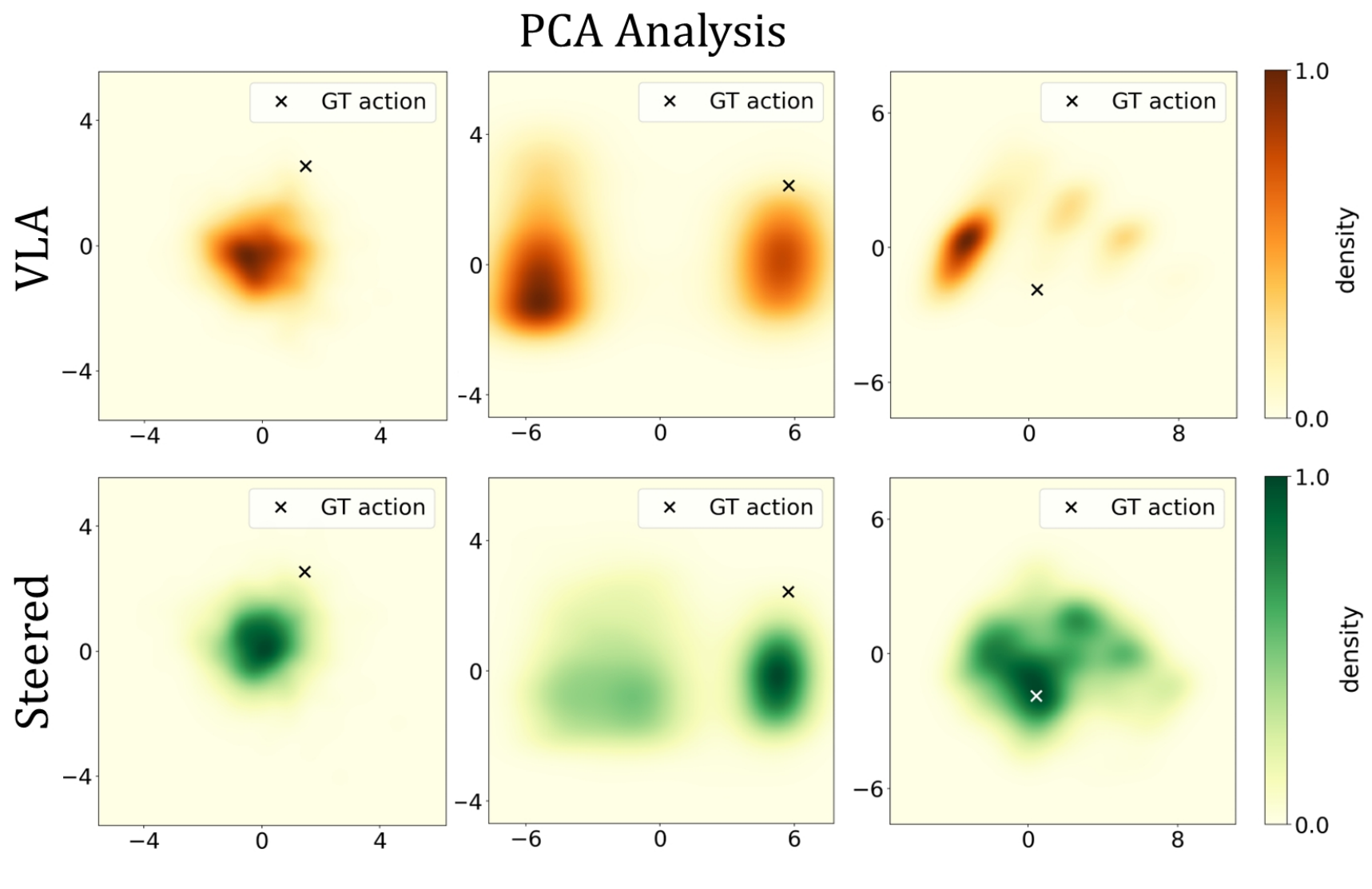}
\vspace{-7mm}
\captionof{figure}{\textbf{PCA Heatmap Analysis:} We represent VLA actions along its two principal axes via Principal Component Analysis on tuples from BridgeV2 dataset~\cite{walke2023bridgedata}. During failure, $RL^2$ will direct action distribution (green) to be closer to the ground truth action.}
\label{fig:rl2_viz_heatmaps}
\end{minipage}
\vspace{-5mm}
\end{figure*}

\vspace{-0.1cm}
\subsection{Additional Studies}
We perform an additional series of experiments to validate the scalability and practicality of our $RL^2$ framework.

\vspace{0.2em}
\subsubsection{\textbf{$RL^2$ scales well with increasing quantity and diversity of samples}}
From our NRMSE analysis in Sec.~\ref{sec:nrmse_inference_scaling_law}, we gain the insight that a VLA can achieve the best performance when we scale both the quantity and diversity of samples during failure states. 
We validate this insight by scaling the number of sampled actions and rephrases on the four SIMPLER tasks with OOD prompts using $\pi_0$, with and without compositional steering. 
From Fig.~\ref{fig:scaling_rephrases_and_samples}, we notice two consistent trends that support our findings. 
Increasing the number of samples ($\text{1}\rightarrow\text{40}$) for a given language prompt, and increasing the diversity of samples ($\text{1}\rightarrow\text{8}$ rephrases) for the same total number of samples (40), both improve success rates across all methods.
$RL^2$ outperforms \textit{Repeated} and \textit{Rephrase} by achieving up to +18.7\% in performance gain when both types of scaling are combined, supporting our scaling laws established in Sec.~\ref{sec:nrmse_inference_scaling_law}.

\vspace{0.2em}
\subsubsection{\textbf{Our alpha selection heuristic is effective in proposing the top per-task alphas during test-time}}
\label{sec:alpha_selection_heuristics_results}
In Sec.~\ref{sec:failure_detection_for_adaptive_steering}, we established the importance of designing an efficient heuristic that selects the optimal CP significance level $\alpha$ for each task, in order to introduce the appropriate amount of steering during failure.
To validate the efficiency of our balanced-accuracy heuristic, we study the success rates for $\alpha=\text{0.05}$ to $\alpha=\text{0.50}$ in intervals of 0.05, and plot the performance gains as we pick the top-k $\alpha$ based on our heuristic, where we test $k=\{1,3,5,10\}$.
From Fig.~\ref{fig:alpha_selection_heuristics}, we observe that most of the performance gains, up to 88.5\% relative to maximum performance, can be derived from simply picking the top-3 alphas using our heuristic.
While there may exist a better metric beyond balanced accuracy (e.g., detection timeliness), we leave further improvements for future work.

\vspace{0.2em}
\subsubsection{\textbf{Our steering policy and failure detection module introduce minimal additional latency}}
\label{sec:time_analysis}
While $RL^2$ introduces new modules to the existing verifier-based action selection frameworks, these are lightweight and do not incur substantial latency compared to the baselines themselves.
We illustrate this by measuring the inference time across the different components for the $\pi_0$ setup~\cite{kwok2026scaling} in Table~\ref{tab:inference_time_analysis}.
As we scale the number of action samples, the inference time for our QAM steering policy and SAFE failure detector remains significantly lower compared to the VLA forward pass.
This is mainly because these networks are lightweight and are conditioned on VLA latents, making them compatible with a wide range of verifier-based frameworks at minimal inference costs.

\vspace{0.2em}
\subsubsection{\textbf{Qualitative analysis highlights the efficacy of compositional steering}}
To better understand the underlying reasons for the effectiveness of compositional steering, we visualize action chunks in our $\pi_0$ simulation (Fig.~\ref{fig:rl2_viz_cover_samples}). 
As discussed in-depth in Sec.~\ref{sec:sim_expts_rephrases} and~\ref{sec:sim_expts_ood}, the VLA samples tend to be directed away from the task object or towards other distractors when the VLA is failing, causing long-term oscillatory behaviors. 
Our $RL^2$ framework steers more samples towards the task object, providing diversity for verifier action selection during failure (more illustrations in Appendix Fig.~\ref{fig:sim_spoon_on_towel}--\ref{fig:sim_dino_on_towel}).

To further analyze the effects of compositional steering, we visualize the $\pi_0$ action chunks for failure tuples from the BridgeV2 dataset (defined in Sec.~\ref{sec:nrmse_inference_scaling_law}), visualized along its two principal axes via Principal Component Analysis (PCA).
From Fig.~\ref{fig:rl2_viz_heatmaps}, we observe that the base VLA action distribution is often concentrated further away from the ground truth action. 
Upon steering, we notice that the action distribution shifts much closer to the ground truth action, highlighting the efficacy of compositional steering in correcting failures.

\begin{figure*}[!t]
\centering
\includegraphics[width=\textwidth]{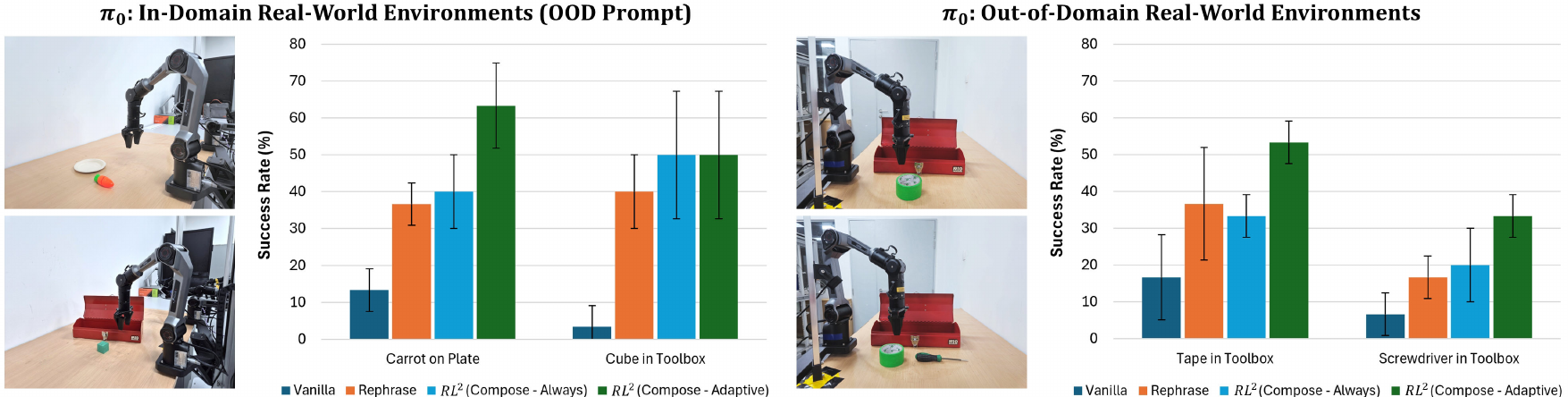}
\vspace{-7mm}
\caption{\textbf{Results of our Real-Robot Experiments:} 
We curate two in-domain and two challenging OOD task environments to compare $RL^2$ against baseline approaches on-hardware.
Adaptive $RL^2$ achieves an average of +17.5\% over \textit{Rephrase}~\cite{kwok2026scaling} baseline, highlighting its robustness to failure modes present in the base VLA.
Similarly, adaptive $RL^2$ achieves an average of +14.2\% over non-adaptive $RL^2$, highlighting the importance of steering only during failures.
}
\vspace{-2mm}
\label{fig:real_robot_pizero_plot}
\end{figure*}

\begin{figure*}[!t]
\centering
\includegraphics[width=\textwidth]{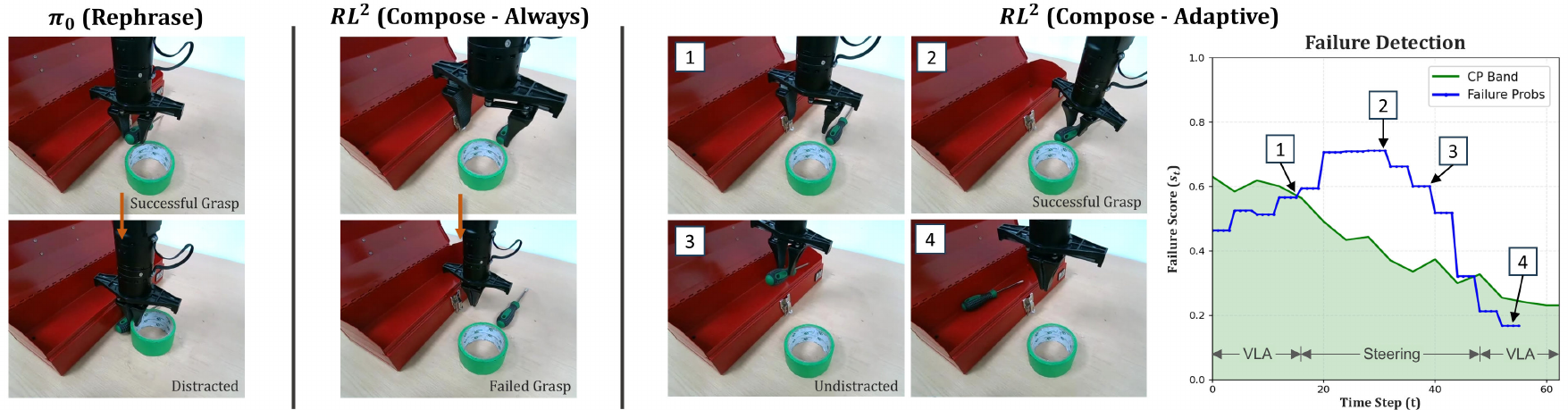}
\vspace{-8.5mm}
\caption{\textbf{Real-Robot Experiment for \textit{Screwdriver in Toolbox} Task (OOD Environment):} This task is challenging for the base VLA ($\pi_0$~\cite{black2026pi0}) because the screwdriver was never seen during training and the scene includes a tape distractor. \textbf{(Left)} \textit{Rephrase} correctly grasps the screwdriver but mistakenly carries the screwdriver to the tape instead of into the box. 
\textbf{(Middle)} Non-adaptive $RL^2$ inaccurately approaches the screwdriver, as it unnecessarily steers already-accurate VLA action samples. 
\textbf{(Right)} Adaptive $RL^2$ relies on accurate VLA samples to approach the screwdriver, then performs compositional steering to guide the samples away from the distractor object and towards the toolbox, when failure is preemptively detected via Conformal Prediction (CP)~\cite{gu2026safe}.
}
\label{fig:real_screwdriver_in_toolbox}
\vspace{-5mm}
\end{figure*}

\vspace{-0.5mm}
\subsection{Real-World Evaluation}
We further validate $RL^2$ on real-world manipulation tasks using a PiperX manipulation arm and a Realsense D405 camera, similar to the WidowX setup used in the BridgeV2 dataset~\cite{walke2023bridgedata}. 
We use the same model weights as our $\pi_0$ simulations, with $\pi_0$~\cite{black2026pi0} as our base VLA, QAM~\cite{li2026qam} as our steering policy, and CoVer~\cite{kwok2026scaling} as our verifier. 
In practice, while these models could generalize well to real-world settings, we found that our SAFE failure detection model~\cite{gu2026safe} could not generalize similarly because it was only trained on a small set of rollouts collected in simulation.
Hence, we attempt to mirror our simulation training setup by collecting real-world rollouts from the four standard SIMPLER in-domain tasks.
We found it challenging to collect rollouts for the \textit{Eggplant in Basket} task because it is difficult to vary the positions of the eggplant given a small sink workspace and a large end-effector (i.e., PiperX has a larger end-effector compared to WidowX). 
Hence, we replaced that task with a new \textit{Cube in Toolbox} task that contains similar objects seen in the BridgeV2 dataset. 

We conduct experiments on four OOD tasks, two of which with red-teaming OOD base prompts and another two with OOD task environments. 
For OOD base prompts, we adopt the \textit{Carrot on Plate} task from SIMPLER and the new \textit{Cube in Toolbox} task.
For OOD task environments, we introduce the \textit{Tape in Toolbox} task where the roll of tape was never seen in the BridgeV2 dataset.
To increase complexity, we also introduce the \textit{Screwdriver in Toolbox} task, where the screwdriver is also unseen and the same tape is included as a distractor.
We repeat all experiments 10 times per seed on 3 random seeds (120 steps), and report our results in Fig.~\ref{fig:real_robot_pizero_plot}.

We find that \textit{Vanilla} $\pi_0$ performs the poorest as it often moves away from or fails to grasp the task object.
While we were able to get better baseline results with \textit{Rephrase}, $RL^2$ with adaptive compositional steering achieves an average performance gain of +17.5\% over \textit{Rephrase}.
Similarly, adaptive $RL^2$ outperforms non-adaptive $RL^2$ by an average of +14.2\%.
We attribute this to the diversity introduced by RL compositional steering during failure states, backed by the following examples.
In Fig.~\ref{fig:real_tape_in_toolbox}, we observe that \textit{Rephrase} outputs samples that often collide with the side of the toolbox, while adaptive $RL^2$ outputs samples that provide sufficient height to lift the tape into the toolbox. 
In Fig.~\ref{fig:real_screwdriver_in_toolbox}, we observe that after the manipulator picked up the screwdriver, \textit{Rephrase} often output action samples that are misguided towards the tape distractor, while adaptive $RL^2$ correctly steers the action samples towards the toolbox.
In appendix Fig.~\ref{fig:real_carrot_on_plate}--\ref{fig:real_cube_in_toolbox}, adaptive $RL^2$ substantially improves VLA replanning. 
For all examples, $RL^2$ initially approaches the task object using the base VLA before SAFE detects failure and triggers compositional steering to avoid or recover from them.
These results align closely with our simulation experiments, and highlight the applicability of the $RL^2$ framework in the real world.

\section{Conclusion}
In this paper, we present $RL^2$, an inference-time steering framework that improves pretrained VLAs through adaptive RL compositional steering conditioned on VLA latents.
We first show that a lightweight RL policy trained on VLA latents and the same offline VLA finetuning dataset can improve downstream performance by steering imitation-learned samples toward diverse actions beyond demonstration modes.
We then discover distinct test-time scaling laws for VLA steering under success and failure states, showing that RL compositional steering is most beneficial when the base policy is likely to fail.
Finally, experiments across VLAs, verifiers, and manipulation benchmarks validate $RL^2$ as a modular framework for in-domain and OOD VLA deployment.
Overall, our results position adaptive test-time steering as a practical pathway toward robust, general-purpose robotics foundation models.
We acknowledge the following limitations of $RL^2$, and leave further improvements to future work: 

\textbf{Steering Function:}
While we evaluated many lightweight steering functions (Sec.~\ref{sec:nrmse_inference_scaling_law}), we have yet to compare our approach with differentiable steering from large models such as VLMs~\cite{liu2026vls} or other VLAs~\cite{lee2025molmoact}.
Since our work focuses on minimizing additional overhead, we leave this to future work. 

\textbf{Failure Detection:}
Our alpha selection heuristic requires additional test-time evaluation to determine the optimal CP significance level $\alpha$, and could be improved with a more effective metric than balanced accuracy (e.g., detection timeliness). 
Furthermore, our failure detection module currently relies on online rollout collection for training. Future work will instead leverage large-scale offline datasets to improve generalization while eliminating the need for data collection.

\textbf{Verifier:} 
Our work assumes a robust verifier that effectively selects among action candidates from $RL^2$.
Future work will relax this assumption by co-training the RL steering policy with the verifier to enhance the robustness of action selection.

\section*{Acknowledgments}
This work was supported by Singapore Technologies Engineering Ltd, under the Economic Development Board - Industrial Postgraduate Program (Project No. 2022-2130).

\bibliographystyle{IEEEtran}
\bibliography{ref}

\clearpage

\begin{center}
    \textsc{Appendix}
\end{center}

\setcounter{section}{0}
\setcounter{subsection}{0}
\setcounter{subsubsection}{0}
\renewcommand{\thesection}{\Alph{section}}
\renewcommand{\thesubsection}{\thesection.\arabic{subsection}}
\renewcommand{\thesubsubsection}{\thesubsection.\arabic{subsubsection}}
\newcommand{\appendixsection}[1]{
    \refstepcounter{section}
    \setcounter{subsection}{0}
    \par\vspace{0.5\baselineskip}
    \noindent\textit{\thesection.\quad #1}\par\nobreak
}
\newcommand{\appendixsubsection}[1]{
    \refstepcounter{subsection}
    \setcounter{subsubsection}{0}
    \par\vspace{0.25\baselineskip}
    \noindent\textit{\arabic{subsection})\quad #1:}~\ignorespaces
}

\appendixsection{Additional Implementation Details}
\label{app:add_implementation_details}
\vspace{1mm}
In this section, we provide additional training details regarding our RL steering policies and failure detection models. 

\appendixsubsection{\textbf{RL Steering Policy}} 
\label{app:steering_policy_details}
The training hyperparameters for QAM~\cite{li2026qam} and V-GPS~\cite{nakamoto2024steering} RL steering policies are listed in Tables~\ref{tab:qam_hyperparameters} and~\ref{tab:vgps_hyperparameters} respectively.
We trained all RL policies on two NVIDIA L40 GPUs, with training times largely determined by the embedding size from the VLA backbones. 

Specifically, we train QAM on $\pi_0$~\cite{black2026pi0} latents using BridgeV2~\cite{walke2023bridgedata} dataset for 500k steps in 4.5 hours. 
In addition, we train QAM on $\pi_{0.5}$~\cite{intelligence2025pi05} latents using DROID~\cite{khazatsky2024droid} dataset for 1M steps in 9 hours.
For the DROID training dataset, we follow the setup in PolaRiS~\cite{jain2025polaris} that uses 90\% DROID data and 10\% simulation data. 
As much as possible, we use the default hyperparameters from the respective repositories.

\begin{table}[htbp]
    \vspace{-0.1cm}
    \small
    \centering
    \caption{QAM~\cite{li2026qam} RL Training hyperparameters.}
    \begin{tabular}{ll}
        \toprule
        \textbf{Hyperparameter} & \textbf{Value} \\
        \midrule
        Batch size & 256 \\
        Discount factor ($\gamma$) & 0.99 \\
        Learning rate & $3 \times 10^{-4}$ \\
        Target network update rate ($\lambda$) & $5 \times 10^{-3}$ \\
        Critic ensemble size ($K$) & 10 \\
        Critic target pessimistic coefficient ($\rho$) & 0.5 \\
        UTD ratio & 1 \\
        Number of flow steps ($T$) & 10 \\
        Network width & 512 \\
        Network depth & 4 hidden layers \\
        Positive reward steps $H$ & 3 \\
        Embedding Dimension ($\pi_{0/0.5}$) & 1024 \\
        Training steps ($\pi_0$ / $\pi_{0.5}$) & $5\times10^5$ / $1\times10^6$ \\        
        Horizon length ($\pi_0$ / $\pi_{0.5}$) & 4 / 15 \\
        Inverse temperature ($\pi_0$ / $\pi_{0.5}$) & 0.1 / 0.02 \\
        \bottomrule
    \end{tabular}
    \vspace{0.5em}
    \vspace{-0.1cm}
    \label{tab:qam_hyperparameters}
\end{table}

In addition, we train V-GPS~\cite{nakamoto2024steering} on OpenVLA latents using BridgeV2~\cite{walke2023bridgedata} dataset for 500k steps (48 hours). 
The training time is longer because the latents dimension used in OpenVLA is four times larger (4096 vs 1024).
We use \textit{Conservative Q-Learning} (CQL)~\cite{aviral2020cql} instead of the default \textit{Calibrated Q-Learning} (CalQL)~\cite{nakamoto2024steering} for training as we empirically find that it performs better when trained on BridgeV2.

\begin{table}[htbp]
    \vspace{-0.1cm}
    \small
    \centering
    \caption{V-GPS~\cite{nakamoto2024steering} RL Training hyperparameters.}
    \begin{tabular}{ll}
        \toprule
        \textbf{Hyperparameter} & \textbf{Value} \\
        \midrule
        CQL $\alpha$ & 1.0 \\
        Discount factor & 0.98 \\
        Learning rate & 3e-4 \\
        Positive reward steps $H$ & 3 \\
        Softmax temperature $\beta$ & 1.0 \\
        Embedding Dimension (OpenVLA) & 4096 \\
        Training steps (OpenVLA) & $5\times10^5$ \\
        \bottomrule
    \end{tabular}
    \vspace{0.5em}
    \vspace{-0.1cm}
    \label{tab:vgps_hyperparameters}
\end{table}

\appendixsubsection{\textbf{VLA Feature Extraction}}
\label{app:vla_feature_extraction}
We borrow different techniques of feature aggregation from SAFE~\cite{gu2026safe} to generate embeddings used by our RL steering policy and SAFE. 
Given the VLA internal feature vector $E \in \mathbb{R}^{n \times d'}$, we aggregate it into a single fixed-dimensional embedding $e \in \mathbb{R}^{d}$, where $n$ can correspond to any dimension (e.g., token position, diffusion steps etc.) and $d$ denotes the feature dimension. 
There are four methods of aggregation proposed by SAFE: 
\begin{itemize}
    \item \textbf{First:} take the first dimension of $E$ along dimension $n$.
    \item \textbf{Last:} take the last dimension of $E$ along dimension $n$.
    \item \textbf{Mean:} take the average of $E$ along dimension $n$.
    \item \textbf{Concat:} concatenate the first and last dimensions of $E$ along dimension $n$.
\end{itemize}

For autoregressive VLAs like OpenVLA~\cite{kim2024openvla}, we extract the feature vector right before it gets decoded into output tokens from the last transformer block. 
For flow-matching VLAs like $\pi_{0/0.5}$~\cite{black2026pi0,intelligence2025pi05}, we extract the feature vectors right before they are used to generate the velocity field.

\appendixsubsection{\textbf{SAFE Failure Detection}}
\label{{app:failure_detection}}
We use the LSTM variant of SAFE, and find the best hyperparameter for SAFE training using grid search that maximizes the ROC-AUC failure detection metric, averaged across 3 seeds.
The optimal hyperparameters used for each of the VLAs are as follows:

\begin{table}[H]
    \vspace{-0.1cm}
    \small
    \centering
    \caption{SAFE~\cite{gu2026safe} Training hyperparameters.}
    \begin{tabular}{ll}
        \toprule
        \textbf{Hyperparameter} & \textbf{Value} \\
        \midrule
        \multicolumn{2}{c}{OpenVLA~\cite{kim2024openvla} (simulation)} \\
        \multicolumn{2}{c}{ROC-AUC: 80.9\%} \\
        \midrule
        Aggregation Method (Token Dim) & Last \\
        Learning rate & 1e-4 \\
        Lambda Regularizer $\lambda_{reg}$ & 1.0 \\
        Batch Size & 64 \\
        Epochs & 1000 \\
        Embedding Dimension & 4096 \\
        \midrule
        \multicolumn{2}{c}{$\pi_{0.5}$~\cite{intelligence2025pi05} (simulation)} \\
        \multicolumn{2}{c}{ROC-AUC: 82.0\%} \\
        \midrule
        Aggregation Method (Horizon Dim) & Mean \\
        Aggregation Method (Diffusion Dim) & First \\
        Learning rate & 3e-4 \\
        Lambda Regularizer $\lambda_{reg}$ & 1e-2 \\
        Batch Size & 512 \\
        Epochs & 1000 \\
        Embedding Dimension & 1024 \\
        \midrule
        \multicolumn{2}{c}{$\pi_0$~\cite{black2026pi0} (simulation)} \\
        \multicolumn{2}{c}{ROC-AUC: 90.4\%} \\
        \midrule
        Aggregation Method (Horizon Dim) & Mean \\
        Aggregation Method (Diffusion Dim) & First \\
        Learning rate & 1e-3 \\
        Lambda Regularizer $\lambda_{reg}$ & 1e-1 \\
        Batch Size & 512 \\
        Epochs & 2000 \\
        Embedding Dimension & 1024 \\
        \midrule
        \multicolumn{2}{c}{$\pi_0$~\cite{black2026pi0} (real-world)} \\
        \multicolumn{2}{c}{ROC-AUC: 85.4\%} \\
        \midrule
        Aggregation Method (Horizon Dim) & Mean \\
        Aggregation Method (Diffusion Dim) & First \\
        Learning rate & 1e-3 \\
        Lambda Regularizer $\lambda_{reg}$ & 1e-2 \\
        Batch Size & 512 \\
        Epochs & 1000 \\
        Embedding Dimension & 1024 \\
        \bottomrule

    \end{tabular}
    \vspace{0.5em}
    \vspace{-0.1cm}
    \label{tab:safe_hyperparameters}
\end{table}

\begin{figure*}[!t]
\centering
\includegraphics[width=0.97\textwidth]{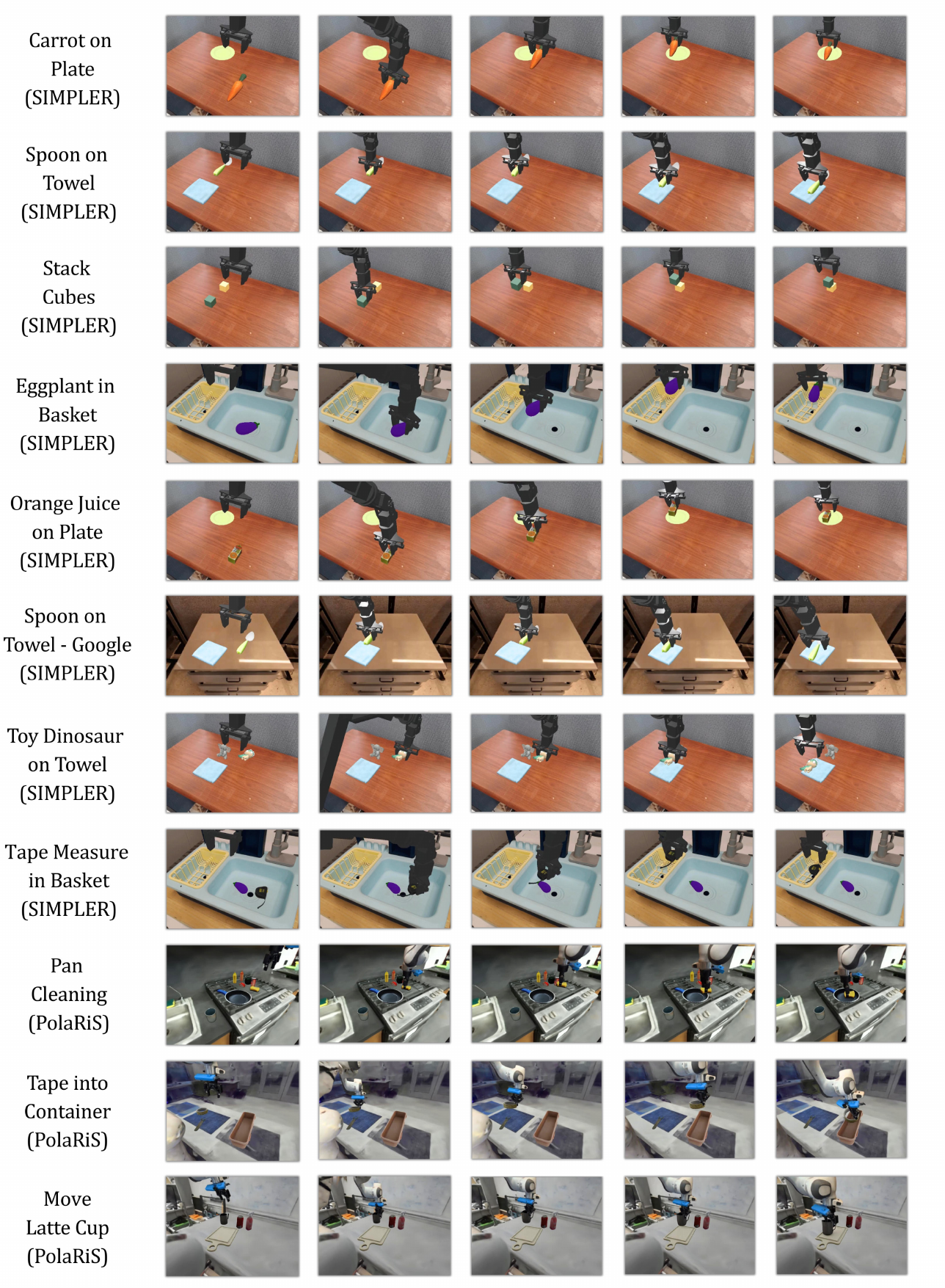}
\vspace{-4mm}
\caption{\textbf{Simulation Tasks:} Task execution for eight SIMPLER tasks~\cite{li24simpler} (WidowX manipulator) and three PolaRiS tasks~\cite{jain2025polaris} (Franka manipulator). }
\label{fig:simpler_polaris_all_tasks}
\end{figure*}

\begin{figure*}[!t]
\centering
\includegraphics[width=0.97\textwidth]{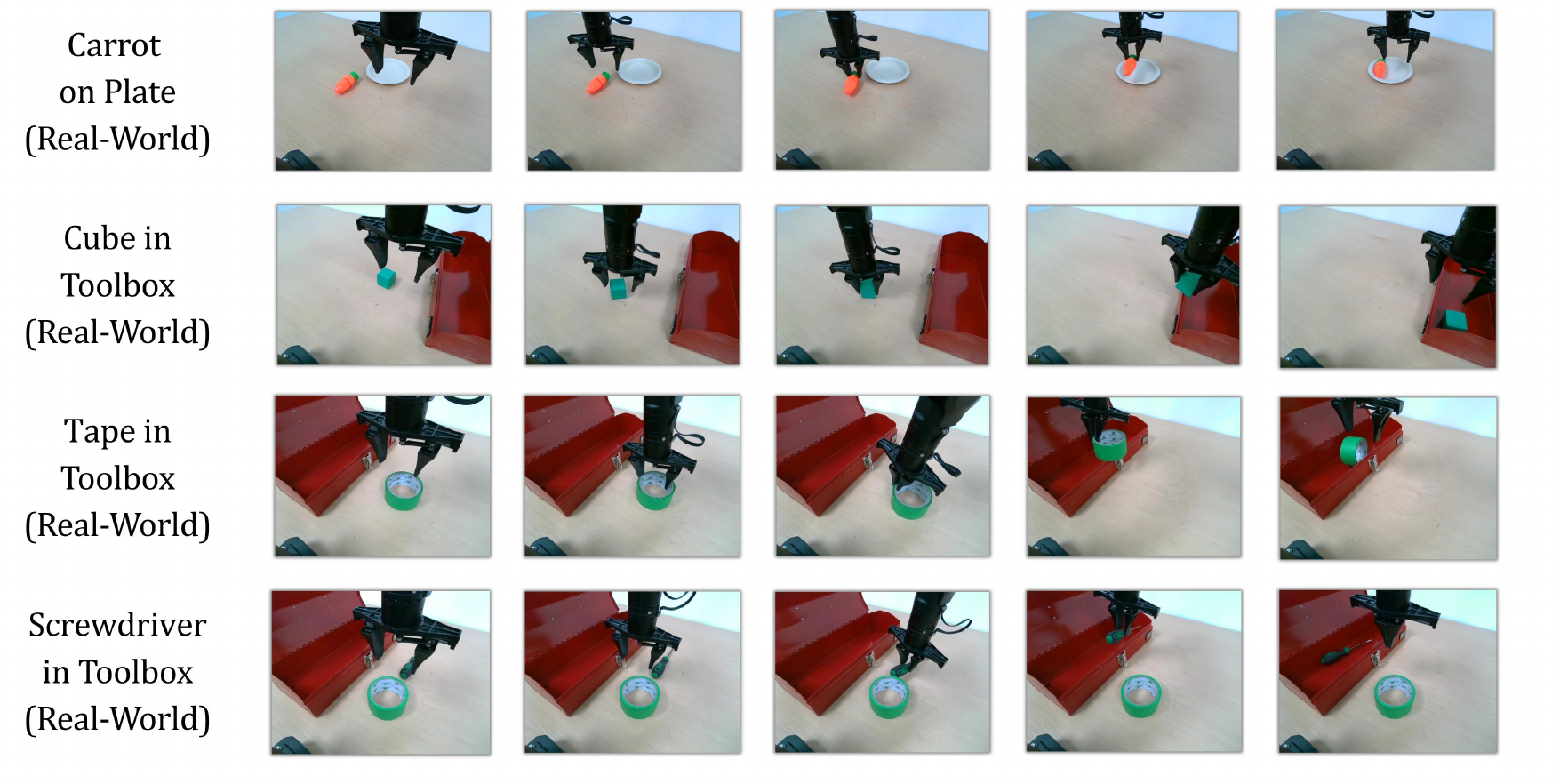}
\vspace{-5mm}
\caption{\textbf{Real-World Tasks:} Task execution for two in-domain and two OOD real-world task environments on the PiperX manipulator.}
\label{fig:real_world_all_tasks}
\end{figure*}

\appendixsection{Additional Task Description}
\label{app:additional_task_description}
\vspace{0.75mm}
In this work, we evaluate $RL^2$ against competing baselines across a wide variety of tasks. 
As introduced in Sec.~\ref{sec:experiments}, we evaluate our approach on 4 in-domain SIMPLER tasks, 8 OOD SIMPLER tasks, 3 OOD PolaRiS tasks, and 4 real-world tasks (Fig.~\ref{fig:simpler_polaris_all_tasks} and~\ref{fig:real_world_all_tasks}). 
These OOD tasks involve OOD instructions (i.e., red-teaming base prompts~\cite{Karnik2024EmbodiedRT}) and OOD task setups (e.g., OOD task objects and/or distractors).
Both SIMPLER and real-world tasks use the WidowX setup, while PolaRiS tasks use the Franka Panda manipulator, following the BridgeV2~\cite{walke2023bridgedata} and DROID~\cite{khazatsky2024droid} dataset setups respectively.

\appendixsubsection{\textbf{BridgeV2 In-domain Simulation Tasks}}
We evaluate OpenVLA~\cite{kim2024openvla} on the set of four standard SIMPLER tasks~\cite{li24simpler} with their original task language instructions. 
The intent is to assess whether $RL^2$ can improve the failure modes for these tasks that were seen by the VLA during training.

\appendixsubsection{\textbf{BridgeV2 OOD Simulation Tasks}}
To elicit the failure modes in VLAs, we evaluate $\pi_0$ using the four standard SIMPLER tasks, but with OOD red-teaming base prompts via the implementation in~\cite{kwok2026scaling}. 
In addition, we introduce four additional OOD tasks from Interleave-VLA~\cite{fan2026interleavevla} that contain OOD task objects, distractors, and variations in background (VLM-generated rephrases can be found in Table~\ref{tab:rephrases_appendix} below).

\begin{itemize}
    \item \textbf{Orange Juice on Plate:} This task is similar to the standard SIMPLER \textit{Carrot on Plate} task, but we instead swap out the carrot for an orange juice container that was not seen during training. 
    \item \textbf{Spoon on Towel (Google):} This task is similar to the standard SIMPLER \textit{Spoon on Towel} task, but the background is changed from the standard brown table to the top surface of an office drawer. 
    \item \textbf{Toy Dinosaur on Towel:} This task uses the same tabletop as the standard SIMPLER \textit{Spoon on Towel} task, but the spoon is changed to a toy dinosaur that was not seen during training. There is a toy elephant placed near the toy dinosaur as a distractor, also not seen during training.
    \item \textbf{Tape Measure in Basket:} This task uses the same sink used in the standard SIMPLER \textit{Eggplant in Basket} task, but instead uses a measuring tape as the task object and the eggplant as a distractor. Note that the measuring tape was never seen during training. 
\end{itemize}

\appendixsubsection{\textbf{PolaRiS OOD Simulation Tasks}}
We evaluate $\pi_{0.5}$ on three tasks from the PolaRiS~\cite{jain2025polaris} benchmark, each with OOD base prompts via the implementation in~\cite{kwok2026scaling}.
Compared to the SIMPLER tasks, these tasks tend to be longer horizon and in the presence of many other distractor objects. 
The VLM-generated rephrases used can be found in Table~\ref{tab:rephrases_appendix} below.

\begin{itemize}
    \item \textbf{Pan Cleaning:} This scene is meant to mimic a kitchen stovetop environment. Aside from the sponge and the cooking pan, this scene also contains many other distractors such as two condiment bottles, an empty cup, a coke can, a latte cup, and a piece of sushi. 
    \item \textbf{Tape into Container:} This scene is of a tabletop with a roll of tape, container, and a wrench distractor object. 
    \item \textbf{Move Latte Cup:} This scene is of a tabletop with a latte cup and a cutting board, along with distractor objects such as coke can and a bottle of hand sanitizer.
\end{itemize}

\begin{table}[t]
\centering
\caption{\textbf{Scaling time analysis:} RL steering and failure detection remain efficient with increasing batch sizes (in ms).}
\vspace{-2mm}
\label{tab:inference_time_analysis_all}
\scriptsize
\setlength{\tabcolsep}{3pt}
\resizebox{\columnwidth}{!}{
\begin{tabular}{c|cccc|cccc}
\toprule
\multirow{2}{*}{\makecell{\textbf{Batch}\\\textbf{Size}}} &
\multicolumn{4}{c|}{\textbf{NVIDIA H100}} &
\multicolumn{4}{c}{\textbf{NVIDIA RTX 5090}} \\
\cmidrule(lr){2-5}\cmidrule(lr){6-9}
& \makecell{$\pi_0$\\\cite{black2026pi0}} & \makecell{CoVer\\\cite{kwok2026scaling}} & \makecell{QAM\\\cite{li2026qam}} & \makecell{SAFE\\\cite{gu2026safe}}
& \makecell{$\pi_0$\\\cite{black2026pi0}} & \makecell{CoVer\\\cite{kwok2026scaling}} & \makecell{QAM\\\cite{li2026qam}} & \makecell{SAFE\\\cite{gu2026safe}} \\
\midrule
1   & 219 & 31 & 9  & 1 & 232  & 96  & 12 & 1 \\
2   & 226 & 32 & 12 & 1 & 243  & 100 & 15 & 1 \\
4   & 231 & 34 & 12 & 1 & 265  & 102 & 16 & 1 \\
8   & 240 & 34 & 14 & 1 & 320  & 107 & 17 & 2 \\
16  & 242 & 35 & 14 & 1 & 434  & 111 & 21 & 2 \\
32  & 286 & 35 & 14 & 1 & 724  & 114 & 27 & 2 \\
64  & 406 & 36 & 19 & 1 & 1541 & 116 & 40 & 2 \\
128 & 698 & 45 & 25 & 1 & 3093 & 145 & 48 & 2 \\
\bottomrule
\end{tabular}
}
\vspace{1mm}
\end{table}

\appendixsubsection{\textbf{BridgeV2 Real-world Tasks}}
We evaluate $\pi_0$ on four OOD real-world tasks, two of which
with red-teaming OOD base prompts and the other two with
OOD task environments.
All of these real-world tasks introduce at least some distribution shift such as camera placement, lighting, and background. 
Tasks involving the toolbox require the VLA to output actions that lift the task object sufficiently high in order to drop the object into the toolbox. 
The VLM-generated rephrases used can be found in Table~\ref{tab:rephrases_appendix} below.

\begin{itemize}
    \item \textbf{Carrot on Plate:} This setup is designed to be identical to the standard SIMPLER \textit{Carrot on Plate} task. While similar, there are some minor differences such as the size of the carrot and the color of the plate.
    \item \textbf{Cube in Toolbox:} This setup contains a green cube and a red toolbox, both of which are in-domain task objects present in the BridgeV2 training set. 
    \item \textbf{Tape in Toolbox:} This setup is similar to the \textit{Cube in Toolbox} task, but the cube is swapped out for a green roll of tape that was not seen during training. 
    \item \textbf{Screwdriver in Toolbox:} This setup is similar to the \textit{Tape in Toolbox} task, but it instead uses a screwdriver as the main task object and the same green roll of tape as the distractor. Both the screwdriver and the tape were not seen during training. 
\end{itemize}

\appendixsubsection{\textbf{Choice of Verifier}}
For in-domain tasks, we use the RoboMonkey action verifier~\cite{kwok2025robomonkey}, which uses LLaVA-7B as the backbone VLM, ViT-Large as the vision encoder, and trained via preference learning on an augmented BridgeV2 dataset. 
The last layer of the VLM is modified to an MLP that outputs preference scores, and selects the action candidate with the highest score.
We use the RoboMonkey implementation directly from~\textcolor{magenta}{\textbf{~\url{https://huggingface.co/robomonkey-vla}}}.

For OOD tasks, we use the CoVer action verifier~\cite{kwok2026scaling} which uses SigLIP2 as the text encoder, ViT-Large as the vision encoder, and a 4-layer stacked transformer encoder to encode a 10-step action trajectory history.
The CoVer verifier is trained via contrastive learning between action embeddings and combined text and vision embeddings (concatenated and then projected), on an augmented BridgeV2/DROID dataset augmented with rephrases.
During inference, the verifier selects action candidates with the highest similarity scores to the text-image pair at the current timestep. 
We use the CoVer 1B implementation directly from~\textcolor{magenta}{\textbf{~\url{https://huggingface.co/cover-vla}}}.

\begin{figure*}[!t]
\centering
\includegraphics[width=0.97\textwidth]{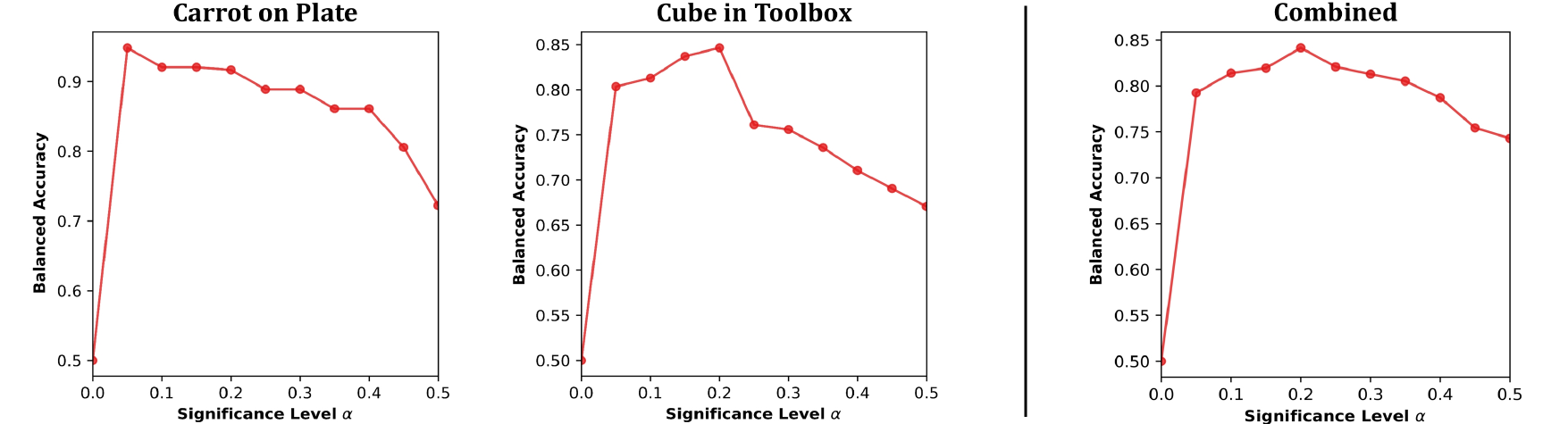}
\vspace{-3mm}
\caption{\textbf{CP Alpha Selection Heuristics:} 
Examples of our Conformal Prediction $\alpha$ selection heuristic comparing balanced accuracy against significance level $\alpha$ plots for real-world tasks, with sample task-wise CP plots (left) and the combined CP plot for OOD tasks (right).
From these plots, we can select the top-k alpha (3, in practice) to evaluate and determine the best CP band to be selected for each task.
}
\vspace{-4mm}
\label{fig:cp_alpha_select_heuristics}
\end{figure*}

\appendixsection{Additional Experimental Details}

\begin{figure*}[!t]
\centering

\includegraphics[width=0.48\textwidth]{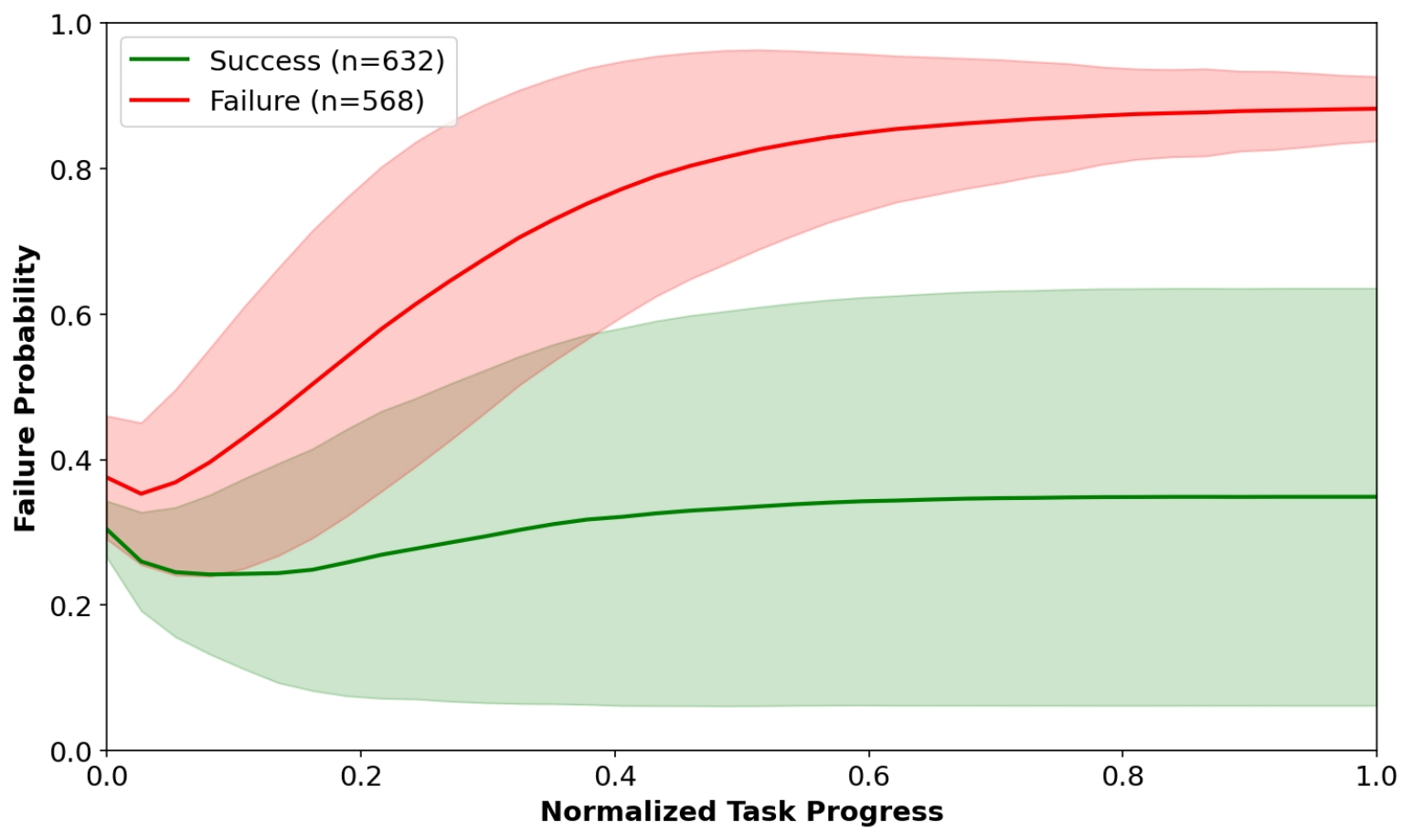}
\hfill
\includegraphics[width=0.48\textwidth]{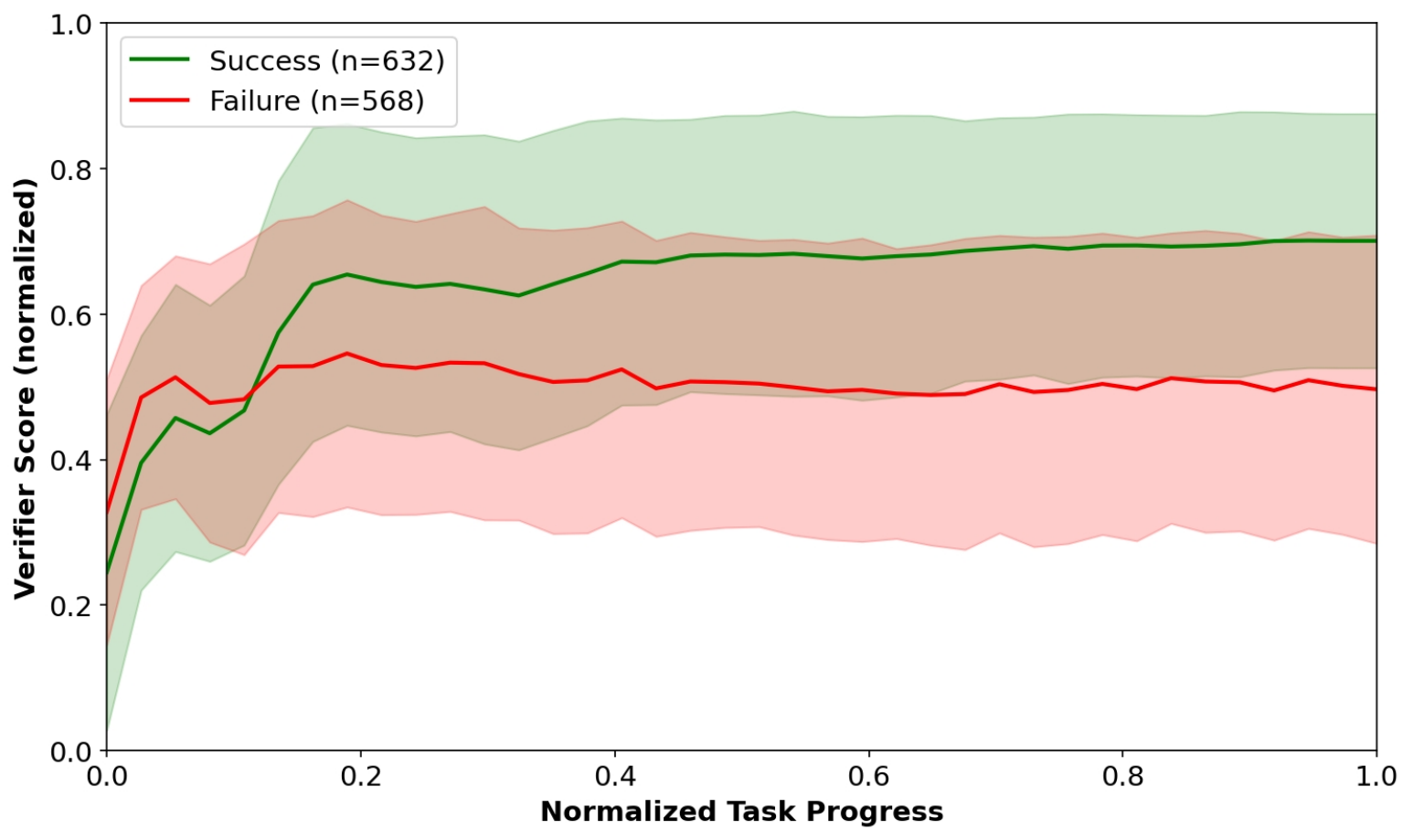}

\vspace{1mm}

\includegraphics[width=0.48\textwidth]{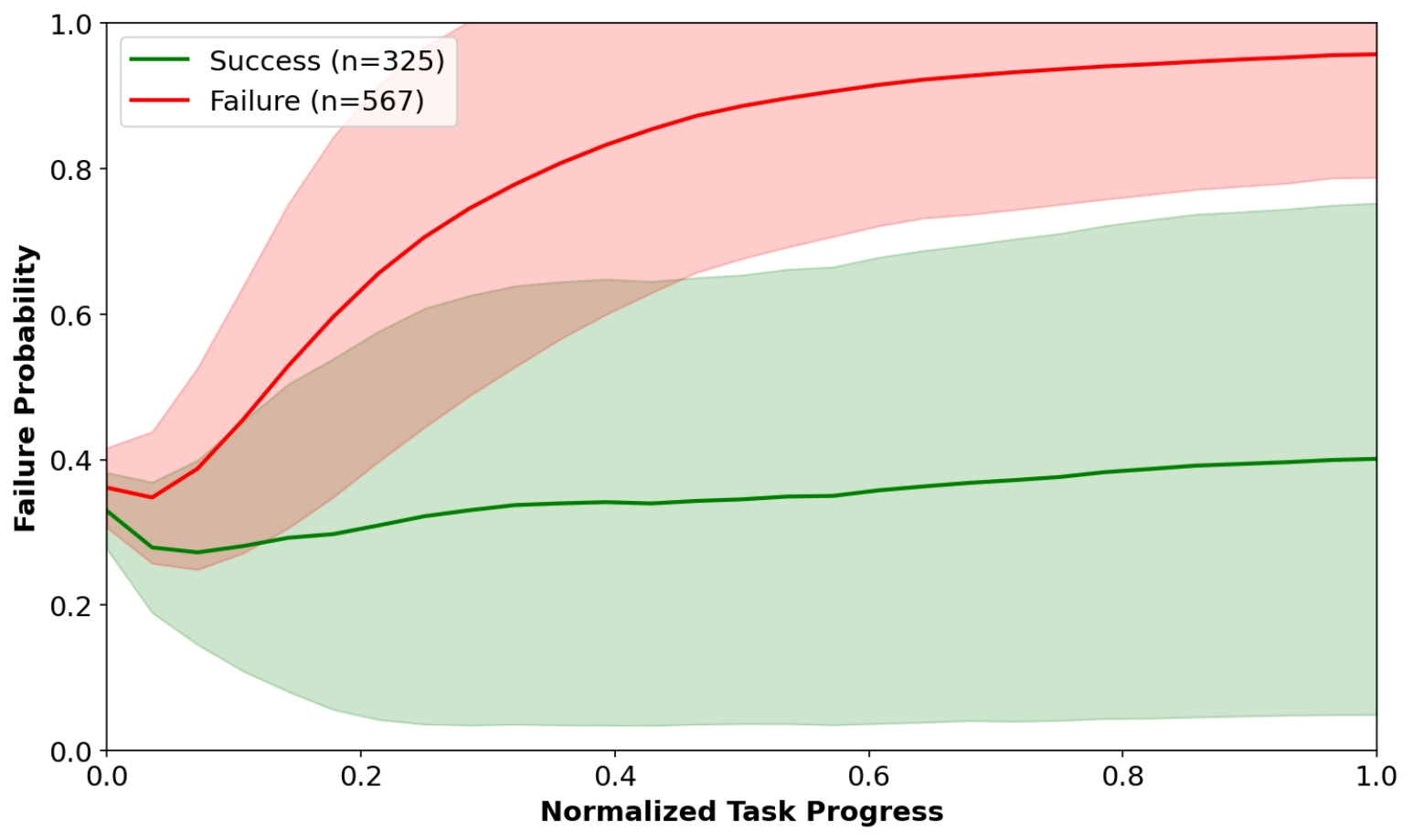}
\hfill
\includegraphics[width=0.48\textwidth]{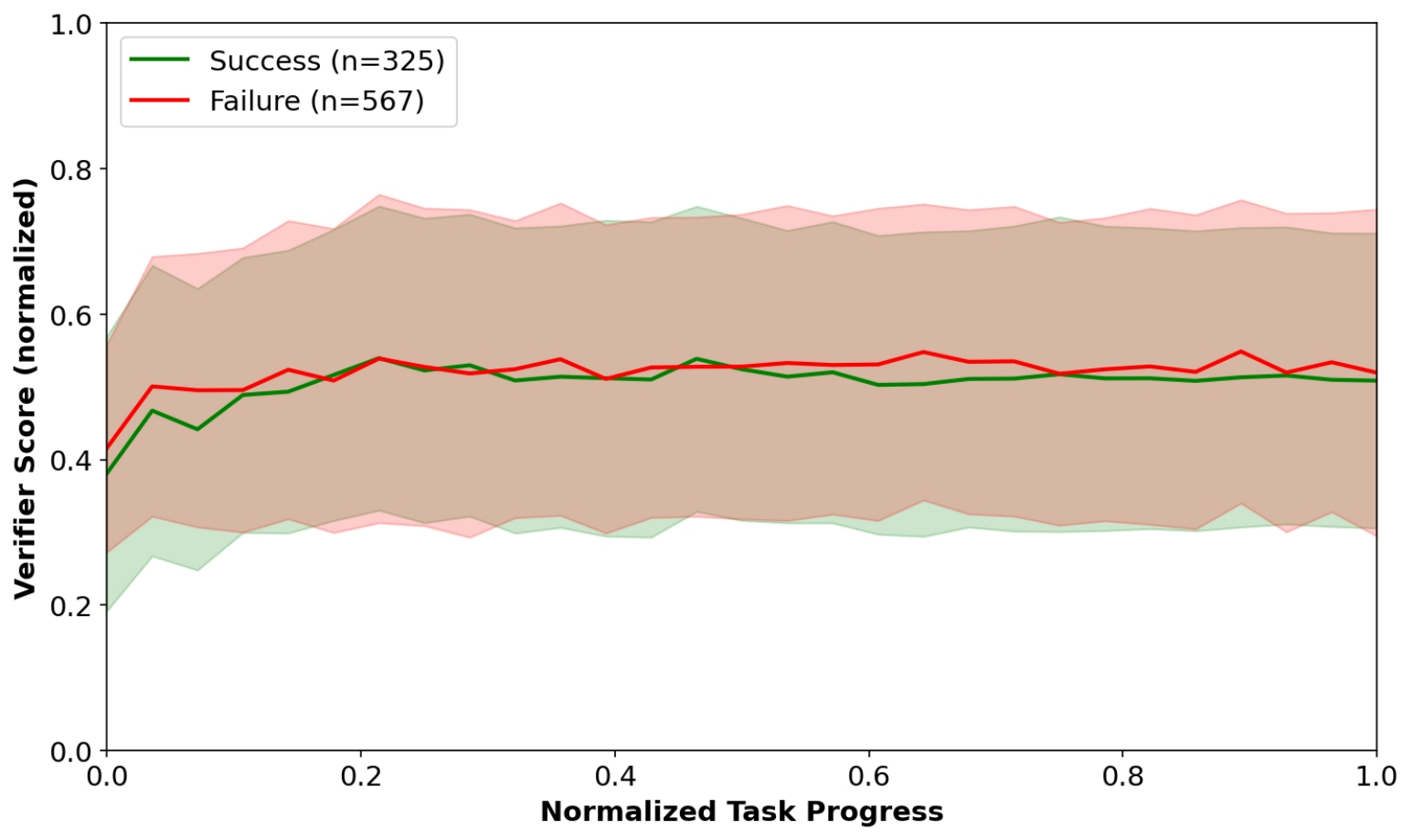}

\vspace{-2mm}
\caption{
\textbf{Comparison of SAFE Failure Probabilities and CoVer Scores:} For both $\pi_0$ and $\pi_{0.5}$ as the base VLA, we visualize SAFE~\cite{gu2026safe} failure probabilities (left) and CoVer~\cite{kwok2026scaling} VLM normalized score (right) against normalized episode progress. 
The solid lines and their corresponding colored bands represent the mean and standard deviation across a large number of episodes $n$.
SAFE acts as an effective intervention trigger because success and failure scores are easily separable, whereas CoVer is less effective as an intervention trigger because success and failure scores are very correlated.
}
\label{fig:safe_vs_cover}
\end{figure*}

\appendixsubsection{\textbf{Intervention Trigger Comparisons}}
\label{app:intervention_comparison}
To further evaluate SAFE~\cite{gu2026safe} as an effective intervention trigger for compositional steering, we plot its average failure probabilities against normalized task progress averaged across a large number of episodes, for both $\pi_0$ and $\pi_{0.5}$.
From Fig.~\ref{fig:safe_vs_cover}, we can see that SAFE is an effective failure detector because the failure probabilities for success and failure rollouts are easily separable.
We further evaluate SAFE by comparing it against our CoVer~\cite{kwok2026scaling} VLM baseline.
As seen from Fig.~\ref{fig:safe_vs_cover}, we observe that the normalized scores for success and failure rollouts are very correlated and hence less separable, rendering it less effective as an intervention trigger compared to SAFE.

\appendixsubsection{\textbf{CP Alpha Selection Heuristics}}
\label{app:alpha_selection_heuristics}
We visualize examples of our Conformal Prediction $\alpha$ selection heuristic by plotting for real-world tasks in Fig.~\ref{fig:cp_alpha_select_heuristics}.
We provide examples for \textit{Carrot on Plate} and \textit{Cube in Toolbox} task-wise plots, and the combined plot generated from rollouts representing all four of our in-domain tasks (described in Sec.~\ref{sec:experiments}). 
From these plots, we can select the top-3 $\alpha$ to evaluate and determine the best CP band for the task (we use combined CP bands for OOD tasks).
When multiple $\alpha$ values are tied in balanced accuracy, we break ties using the relative standalone performance of the RL policy and the VLA: if the RL policy outperforms the VLA, we choose the larger $\alpha$ to permit more frequent steering (vice versa).
This is consistent with the findings in~\cite{cao2026compose} where composition is most effective when biased towards the better performing policy. 
Note that for our OpenVLA and $\pi_{0/0.5}$ simulation experiments, we generate these plots per seed, allowing for more granularity in choosing the best CP band per seed.
Importantly, we only need to perform this alpha selection heuristic once whenever we introduce a new task.

\appendixsubsection{\textbf{Additional Time Analysis}}
Apart from the time analysis for our $RL^2$ framework on an NVIDIA RTX5090 GPU (Sec.~\ref{sec:time_analysis}), we include additional time analysis on an NVIDIA H100 GPU.
From Table~\ref{tab:inference_time_analysis_all}, we observe a similar scaling trend across both hardware, where the latency for our QAM and SAFE modules remains significantly lower compared to the latency due to the VLA itself. 
This highlights the applicability of $RL^2$ especially in real-world deployment.

\begin{figure*}[!t]
\centering
\includegraphics[width=\textwidth]{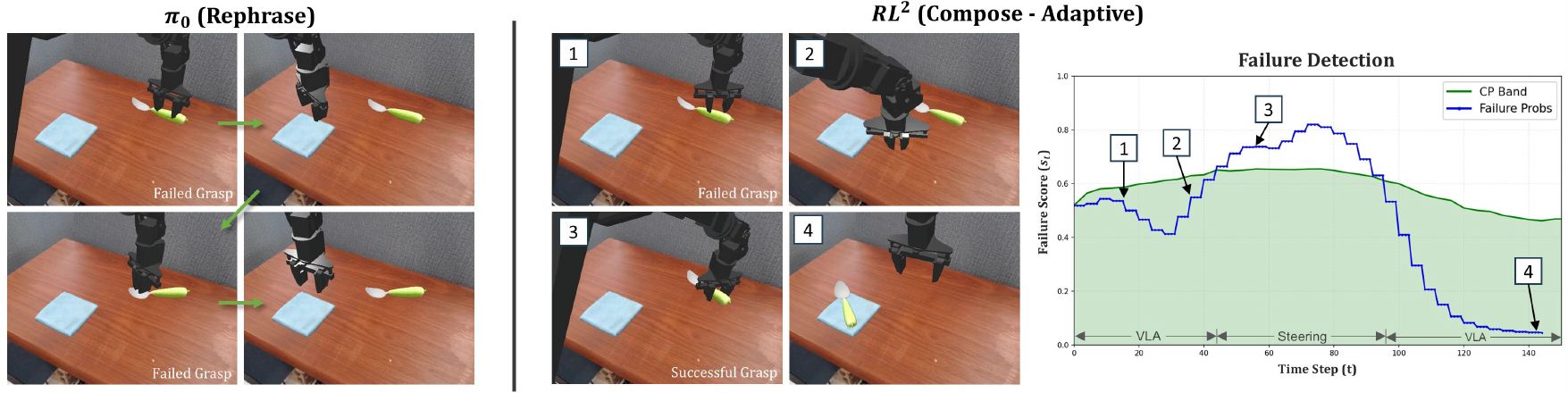}
\vspace{-8mm}
\caption{\textbf{SIMPLER Experiment for \textit{Spoon on Towel} Task (OOD Prompt):} 
This task is challenging for the base VLA ($\pi_0$~\cite{black2026pi0}) because the base language prompt was never seen during training. 
\textbf{(Left)} \textit{Rephrase} repeatedly fails to grasp the spoon  during replanning.
\textbf{(Right)} Adaptive $RL^2$ is more likely to successfully grasp the spoon during replanning, after failure is preemptively detected via Conformal Prediction (CP)~\cite{gu2026safe}.}
\label{fig:sim_spoon_on_towel}
\end{figure*}

\begin{figure*}[!t]
\centering
\includegraphics[width=\textwidth]{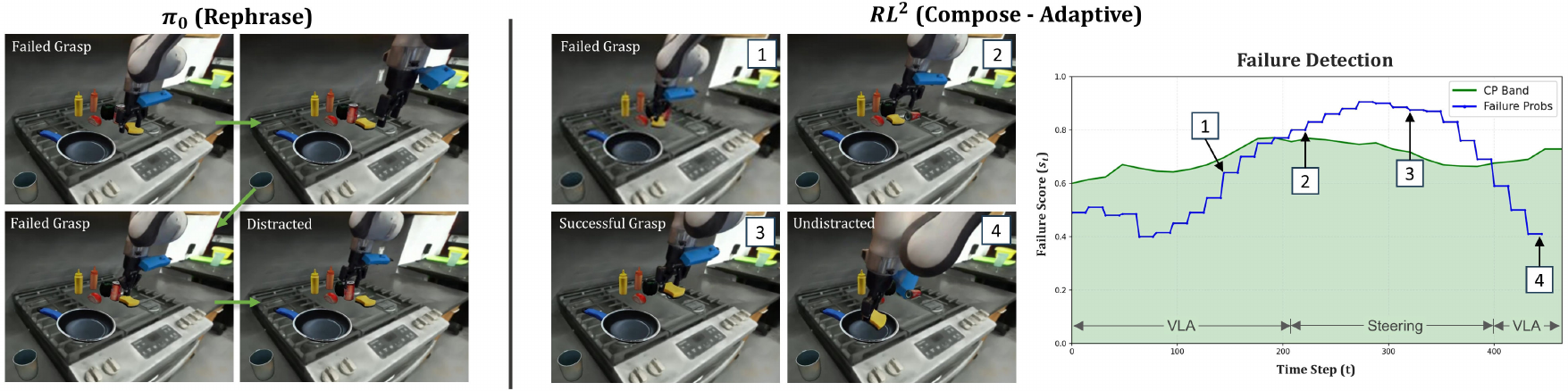}
\vspace{-8mm}
\caption{\textbf{PolaRiS Experiment for \textit{Pan Cleaning} Task (OOD Prompt):} 
This task is challenging for the base VLA ($\pi_0$~\cite{black2026pi0}) because the base language prompt was never seen during training. 
\textbf{(Left)} \textit{Rephrase} repeatedly fails to grasp the sponge  during replanning and eventually gets distracted by the coke can.
\textbf{(Right)} Adaptive $RL^2$ is more likely to successfully grasp the sponge during replanning, after failure is preemptively detected via Conformal Prediction (CP)~\cite{gu2026safe}.}
\label{fig:sim_pan_clean}
\vspace{-0.5cm}
\end{figure*}

\begin{figure*}[!t]
\centering
\includegraphics[width=\textwidth]{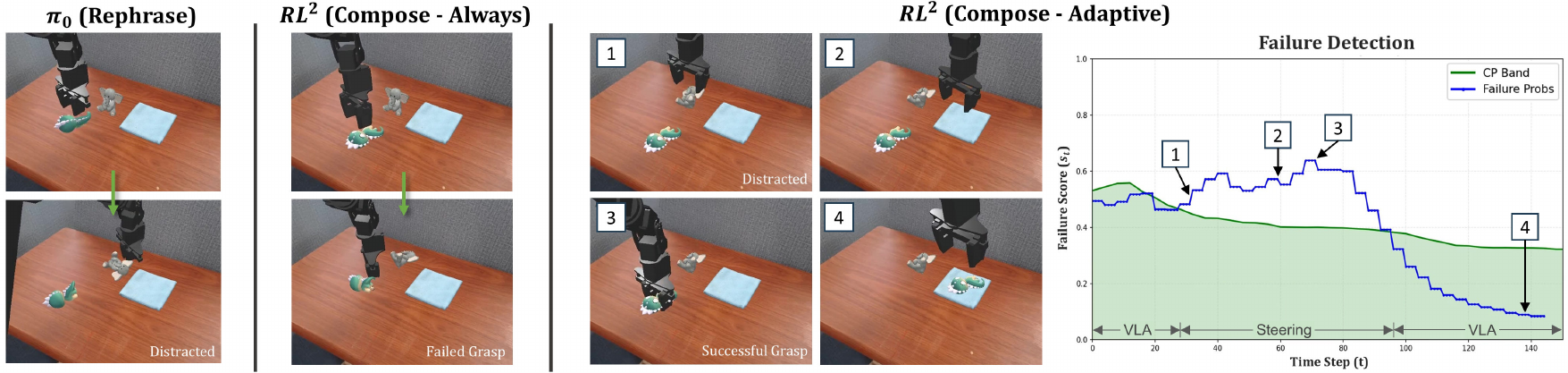}
\vspace{-8mm}
\caption{\textbf{SIMPLER Experiment for \textit{Toy Dinosaur on Towel} Task (OOD Environment):} This task is challenging for the base VLA ($\pi_0$~\cite{black2026pi0}) because the toy dinosaur and elephant were never seen during training. 
\textbf{(Left)} \textit{Rephrase} initially approaches the dinosaur but is eventually distracted by the elephant. 
\textbf{(Middle)} Non-adaptive $RL^2$ consistently approaches the dinosaur but inaccurately attempts to grasp it.
\textbf{(Right)} Adaptive $RL^2$ relies on initially accurate VLA samples to approach the dinosaur, then steers actions to complete the pick-and-place when failure is preemptively detected via Conformal Prediction (CP)~\cite{gu2026safe}.}
\label{fig:sim_dino_on_towel}
\end{figure*}

\begin{figure*}[!t]
\centering
\includegraphics[width=\textwidth]{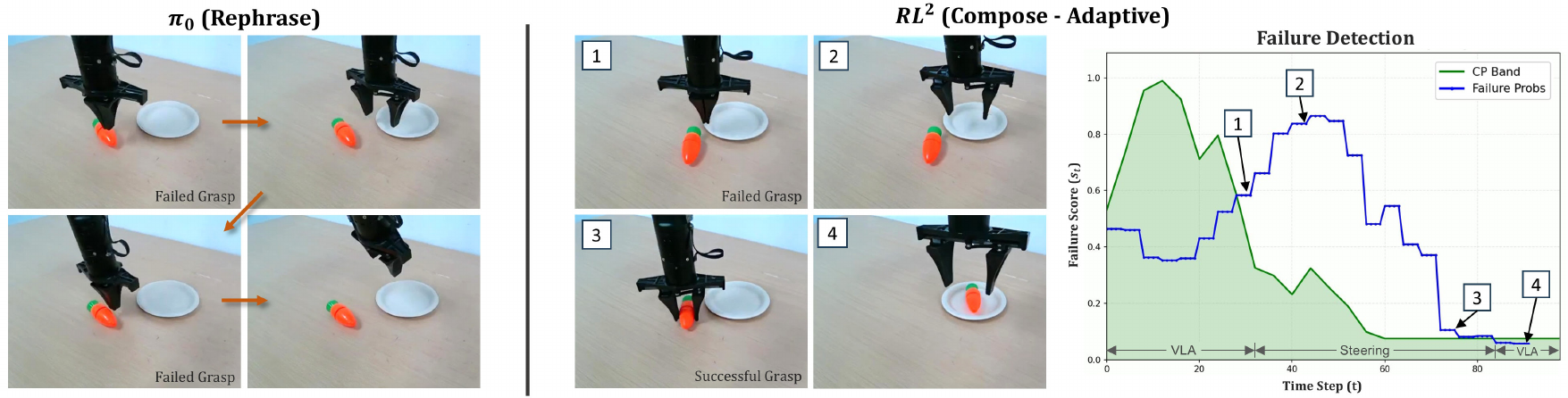}
\vspace{-8mm}
\caption{\textbf{Real-Robot Experiment for \textit{Carrot on Plate} Task (OOD Prompt):} 
This task is challenging for the base VLA ($\pi_0$~\cite{black2026pi0}) because the base language prompt was never seen during training. 
\textbf{(Left)} \textit{Rephrase} repeatedly fails to grasp the carrot  during replanning.
\textbf{(Right)} Adaptive $RL^2$ is more likely to successfully grasp the carrot during replanning, after failure is preemptively detected via Conformal Prediction (CP)~\cite{gu2026safe}.}
\label{fig:real_carrot_on_plate}
\end{figure*}

\begin{figure*}[!t]
\centering
\includegraphics[width=\textwidth]{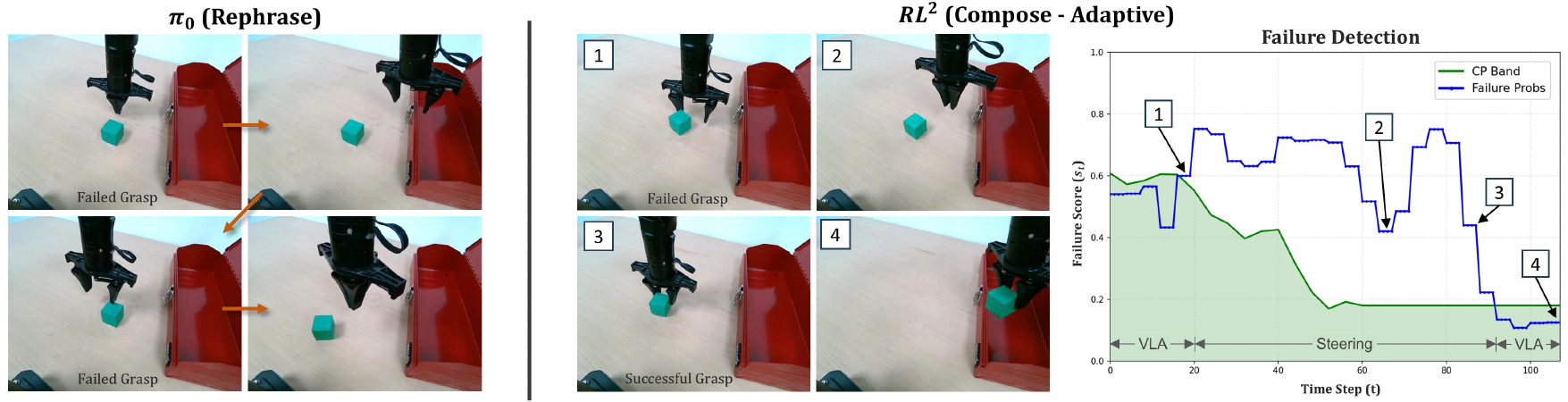}
\vspace{-8mm}
\caption{\textbf{Real-Robot Experiment for \textit{Cube in Toolbox} Task (OOD Prompt):} 
This task is challenging for the base VLA ($\pi_0$~\cite{black2026pi0}) because the base language prompt was never seen during training. 
\textbf{(Left)} \textit{Rephrase} repeatedly fails to grasp the cube  during replanning.
\textbf{(Right)} Adaptive $RL^2$ is more likely to successfully grasp the cube during replanning, after failure is preemptively detected via Conformal Prediction (CP)~\cite{gu2026safe}.}
\label{fig:real_cube_in_toolbox}
\end{figure*}

\begin{table*}[!t]
\centering
\caption{\textbf{Language Instructions Rephrases:} The table lists the original and red-team instructions, along with VLM-generated rephrases across all benchmarks following~\cite{kwok2026scaling}. We provide the red-team instructions only for tasks using OOD base prompts.}
\vspace{-2mm}
\label{tab:rephrases_appendix}
\small
\setlength{\tabcolsep}{4pt}
\renewcommand{\arraystretch}{1.05}
\begin{tabular}{@{}L{0.10\textwidth}L{0.18\textwidth}L{0.18\textwidth}L{0.49\textwidth}@{}}
\toprule
\textbf{Task Name} & \textbf{Original Instruction} & \textbf{Red-Team Instruction} & \textbf{Generated Rephrases} \\
\midrule
Carrot on Plate\newline\textit{\textbf{(SIMPLER})} &
Put carrot on plate. &
Balance the carrot on the ceramic platter. &
1. Place the carrot on the green dish.\newline
2. Set the orange vegetable on the small plate.\newline
3. Position the carrot upright on the platter.\newline
4. Stand the carrot on the green plate.\newline
5. Set the vegetable on the round dish.\newline
6. Position the carrot on the small platter.\newline
7. Balance the orange carrot on the green dish.\newline
8. Place the vegetable on the ceramic plate. \\
\midrule
\midrule
\multicolumn{4}{@{}r@{}}{\textit{Continued on next page}}\\
\end{tabular}
\end{table*}

\begin{table*}[!t]
\addtocounter{table}{-1}
\centering
\caption[]{\textbf{Language Instructions Rephrases} (continued from previous page).}
\vspace{-2mm}
\label{tab:rephrases_appendix}
\small
\setlength{\tabcolsep}{4pt}
\renewcommand{\arraystretch}{1.05}
\begin{tabular}{@{}L{0.10\textwidth}L{0.18\textwidth}L{0.18\textwidth}L{0.49\textwidth}@{}}
\toprule
\textbf{Task Name} & \textbf{Original Instruction} & \textbf{Red-Team Instruction} & \textbf{Generated Rephrases} \\
\midrule
Spoon on Towel\newline\textit{\textbf{(SIMPLER})} &
Put the spoon on the towel. &
Set the spoon exactly in the middle of the towel. &
1. Place the spoon in the center of the towel.\newline
2. Put the utensil on the middle of the cloth.\newline
3. Position the scoop at the center of the fabric.\newline
4. Lay the green-handled spoon in the middle of the blue towel.\newline
5. Set the utensil in the central part of the towel.\newline
6. Position the spoon on the center of the blue towel.\newline
7. Lay the green-handled spoon in the middle of the fabric.\newline
8. Set the spoon in the center of the towel. \\
\midrule
Stack Cubes\newline\textit{\textbf{(SIMPLER})} &
Stack the green block on the yellow block. &
Arrange the lush green element atop the yellowish-orange element. &
1. Place the green cube on the yellow block.\newline
2. Put the grassy block on top of the yellow piece.\newline
3. Set the green block over the yellow cube.\newline
4. Stack the green piece onto the yellowish block.\newline
5. Position the lush cube above the yellow element.\newline
6. Place the green cube on the yellow cube.\newline
7. Set the green block on top of the yellow block.\newline
8. Put the green element over the yellow cube. \\
\midrule
Eggplant in Basket\newline\textit{\textbf{(SIMPLER})} &
Put eggplant into yellow basket. &
Arrange the eggplant neatly in the yellow bin. &
1. Place the vegetable in the yellow rack.\newline
2. Put the eggplant in the yellow holder.\newline
3. Set the purple vegetable into the yellow bin.\newline
4. Organize the eggplant in the yellow rack.\newline
5. Position the vegetable in the yellow holder.\newline
6. Place the purple vegetable in the yellow container.\newline
7. Set the eggplant into the dish rack.\newline
8. Move the purple produce to the yellow bin. \\
\midrule
Orange Juice on Plate\newline\textit{\textbf{(SIMPLER})} &
Put orange juice on plate. &
-- &
1. Set the carton of juice onto the dish.\newline
2. Move the orange drink to the saucer.\newline
3. Lay the juice carton on the platter.\newline
4. Position the orange juice on the dish.\newline
5. Put the drink carton on the yellow plate.\newline
6. Place the orange drink container onto the platter.\newline
7. Set the juice on the yellow saucer.\newline
8. Move the orange box to the plate. \\
\midrule
Spoon on Towel (Google)\newline\textit{\textbf{(SIMPLER})} &
Put the spoon on the towel google. &
-- &
1. Place the spoon in the center of the towel.\newline
2. Put the utensil on the middle of the cloth.\newline
3. Position the scoop at the center of the fabric.\newline
4. Lay the green-handled spoon in the middle of the blue towel.\newline
5. Set the utensil in the central part of the towel.\newline
6. Position the spoon on the center of the blue towel.\newline
7. Set the spoon exactly in the middle of the towel.\newline
8. Set the spoon in the center of the towel. \\
\midrule
Toy Dinosaur on Towel\newline\textit{\textbf{(SIMPLER})} &
Put the toy dinosaur on the towel. &
-- &
1. Place the toy on the cloth.\newline
2. Move the figure to the blue fabric.\newline
3. Transfer the creature to the piece of cloth.\newline
4. Lay the toy on the rag.\newline
5. Position the dinosaur on the towel.\newline
6. Put the object on the blue fabric.\newline
7. Set the figure on the table cloth.\newline
8. Move the dinosaur over to the rag. \\
\midrule
Tape Measure in Basket\newline\textit{\textbf{(SIMPLER})} &
Put tape measure into yellow basket. &
-- &
1. Move the measuring tape to the yellow basket.\newline
2. Set the tape measure inside the yellow basket.\newline
3. Transfer the measuring tape to the yellow container.\newline
4. Put the tape measure in the yellow dish rack.\newline
5. Place the measuring tape into the yellow holder.\newline
6. Place the tape measure in the yellow rack.\newline
7. Set the measuring tape in the yellow rack.\newline
8. Position the tape measure in the yellow basket. \\
\midrule
\midrule
\multicolumn{4}{@{}r@{}}{\textit{Continued on next page}}\\
\end{tabular}
\end{table*}

\makeatletter
\setlength{\@fptop}{0pt}
\setlength{\@dblfptop}{0pt}
\makeatother

\begin{table*}[!t]
\addtocounter{table}{-1}
\centering
\caption[]{\textbf{Language Instructions Rephrases} (continued from previous page).}
\vspace{-2mm}
\small
\setlength{\tabcolsep}{4pt}
\renewcommand{\arraystretch}{1.05}
\begin{tabular}{@{}L{0.10\textwidth}L{0.18\textwidth}L{0.18\textwidth}L{0.49\textwidth}@{}}
\toprule
\textbf{Task Name} & \textbf{Original Instruction} & \textbf{Red-Team Instruction} & \textbf{Generated Rephrases} \\
\midrule
Pan Cleaning\newline\textit{\textbf{(PolaRiS})} &
Use the yellow sponge to scrub the blue handle frying pan. &
Make contact between the cleaning sponge and the frying pan with the blue handle to scrub it. &
1. Scrub the blue-handled frying pan with the sponge.\newline
2. Use the cleaning sponge to clean the pan with the blue handle.\newline
3. Rub the sponge against the frying pan to scrub it.\newline
4. Clean the skillet with the blue handle using the sponge.\newline
5. Touch the cleaning sponge to the blue-handled pan.\newline
6. Wipe the frying pan with the sponge.\newline
7. Apply the sponge to the pan with the blue handle for scrubbing.\newline
8. Wash the blue-handled pan using the cleaning sponge. \\
\midrule
Tape into Container\newline\textit{\textbf{(PolaRiS})} &
Put the tape into the container. &
Carefully drop the adhesive roll inside the bin. &
1. Place the tape roll in the box.\newline
2. Drop the masking tape into the tray.\newline
3. Move the tape to the brown container.\newline
4. Set the roll of tape inside the bin.\newline
5. Lift the tape and place it in the tray.\newline
6. Insert the adhesive tape into the box.\newline
7. Deposit the tape roll inside the receptacle.\newline
8. Grab the tape and put it into the container.\\
\midrule
Move Latte Cup\newline\textit{\textbf{(PolaRiS})} &
Put the latte art cup on top of the cutting board. &
Relocate the pitcher with the decoration onto the wooden board. &
1. Place the metal pitcher on the cutting board.\newline
2. Move the latte cup onto the wooden board.\newline
3. Set the silver cup on the cutting board.\newline
4. Lift the milk jug and put it on the board.\newline
5. Position the frothing pitcher on the wooden board.\newline
6. Take the metal cup and place it on the cutting board.\newline
7. Move the latte pitcher to the board.\newline
8. Put the latte cup on top of the cutting board. \\
\midrule
Carrot on Plate\newline\textit{\textbf{(Real-World})} &
Put carrot on plate. &
Balance the carrot on the ceramic platter. &
1. Place the carrot on the green dish.\newline
2. Set the orange vegetable on the small plate.\newline
3. Position the carrot upright on the platter.\newline
4. Stand the carrot on the green plate.\newline
5. Set the vegetable on the round dish.\newline
6. Position the carrot on the small platter.\newline
7. Balance the orange carrot on the green dish.\newline
8. Place the vegetable on the ceramic plate. \\
\midrule
Cube in Toolbox\newline\textit{\textbf{(Real-World})} &
Place the cube inside the toolbox. &
Move the cube towards the container and put it inside. &
1. Put the cube into the toolbox.\newline
2. Insert the cube into the case.\newline
3. Move the cube into the container.\newline
4. Set the cube inside the red box.\newline
5. Place the cube inside the red box.\newline
6. Place the cube into the box.\newline
7. Insert the cube in the open toolbox.\newline
8. Put the cube inside the case. \\
\midrule
Tape in Toolbox\newline\textit{\textbf{(Real-World})} &
Place the green tape inside the toolbox. &
-- &
1. Put the green tape in the toolbox.\newline
2. Set the tape inside the red box.\newline
3. Move the green roll into the storage box.\newline
4. Insert the green adhesive into the toolbox.\newline
5. Place the colored tape in the red container.\newline
6. Drop the green roll inside the toolbox.\newline
7. Put the adhesive inside the storage box.\newline
8. Set the tape in the container. \\
\midrule
Screwdriver in Toolbox\newline\textit{\textbf{(Real-World})} &
Place the screwdriver tool inside the toolbox. &
-- &
1. Put the screwdriver into the toolbox.\newline
2. Insert the tool into the case.\newline
3. Move the screwdriver into the container.\newline
4. Set the tool inside the red box.\newline
5. Place the tool inside the red box.\newline
6. Place the green and black tool into the box.\newline
7. Insert the screwdriver in the open toolbox.\newline
8. Put the green-handled tool inside the case. \\
\midrule
\end{tabular}
\end{table*}

\vfill

\end{document}